\documentclass[10pt,twocolumn,letterpaper]{article}

\usepackage{iccv}              % To produce the CAMERA-READY version
\definecolor{iccvblue}{rgb}{0.21,0.49,0.74}
\usepackage[pagebackref,breaklinks,colorlinks,allcolors=iccvblue]{hyperref}
\usepackage{times}  % DO NOT CHANGE THIS
\usepackage{helvet}  % DO NOT CHANGE THIS
\usepackage{courier}  % DO NOT CHANGE THIS
\usepackage[hyphens]{url}  % DO NOT CHANGE THIS
\usepackage{graphicx} % DO NOT CHANGE THIS
\usepackage{natbib}  % DO NOT CHANGE THIS AND DO NOT ADD ANY OPTIONS TO IT
\usepackage{caption} % DO NOT CHANGE THIS AND DO NOT ADD ANY OPTIONS TO IT
\usepackage{algorithm}
\usepackage{algorithmic}
\usepackage{microtype}      % microtypography
\usepackage{xcolor}         % colors
\usepackage{amsmath}
\usepackage{multirow}
\usepackage{array}    % 줄 간격 조정을 위해 필요
\usepackage{colortbl}
\usepackage{amssymb} % 추가된 패키지
\usepackage{pifont} % 추가된 패키지
\usepackage{booktabs}
\usepackage{makecell}
\usepackage{footmisc}          % 각주 기호 변경을 쉽게 해주는 패키지

\definecolor{LightBlue}{rgb}{0.85 0.92 0.96}
\definecolor{custom_gray}{gray}{.92}

\definecolor{darkgreen}{RGB}{0,130,0}
\definecolor{darkred}{RGB}{180,0,0}

\definecolor{cellcol}{gray}{.92}

\def\paperID{7296} % *** Enter the Paper ID here
\def\confName{ICCV}
\def\confYear{2025}

\title{MemoryTalker: Personalized Speech-Driven 3D Facial Animation \\via Audio-Guided Stylization}

\author{
  Hyung Kyu Kim\textsuperscript{1} \qquad
  Sangmin Lee\textsuperscript{2,}\thanks{Corresponding Author} \qquad
  Hak Gu Kim\textsuperscript{1,}\footnotemark[2]
  \\[1.5ex]
  \textsuperscript{1}Chung-Ang University, \textsuperscript{2}Korea University\\
  {\tt\small \{hyung1208,hakgukim\}@cau.ac.kr, sangmin-lee@korea.ac.kr}
}

\begin{document}
\maketitle
%=========================== ABSTRACT ===========================
\begin{abstract}
Speech-driven 3D facial animation aims to synthesize realistic facial motion sequences from given audio, matching the speaker’s speaking style. 
However, previous works often require priors such as class labels of a speaker or additional 3D facial meshes at inference, which makes them fail to reflect the speaking style and limits their practical use. 
To address these issues, we propose \textit{MemoryTalker} which enables realistic and accurate 3D facial motion synthesis by reflecting {speaking style} only with audio input to maximize usability in applications. 
Our framework consists of two training stages: $<$1-stage$>$ is storing and retrieving general motion (\textit{i.e.}, Memorizing), and $<$2-stage$>$ is to perform the personalized facial motion synthesis (\textit{i.e.}, Animating) with the motion memory stylized by the audio-driven speaking style feature. 
In this second stage, our model learns about which facial motion types should be emphasized for a particular piece of audio. As a result, our \textit{MemoryTalker} can generate a reliable personalized facial animation without additional prior information. 
With quantitative and qualitative evaluations, as well as user study, we show the effectiveness of our model and its performance enhancement for personalized facial animation over state-of-the-art methods.
Project page: \url{https://cau-irislab.github.io/ICCV25-MemoryTalker/}
\end{abstract}
%=================================================================================

%------------------------------------ Fig_1
%##################################################################################################
\begin{figure}[!t]
\begin{center}
\includegraphics[width=1.0\linewidth]{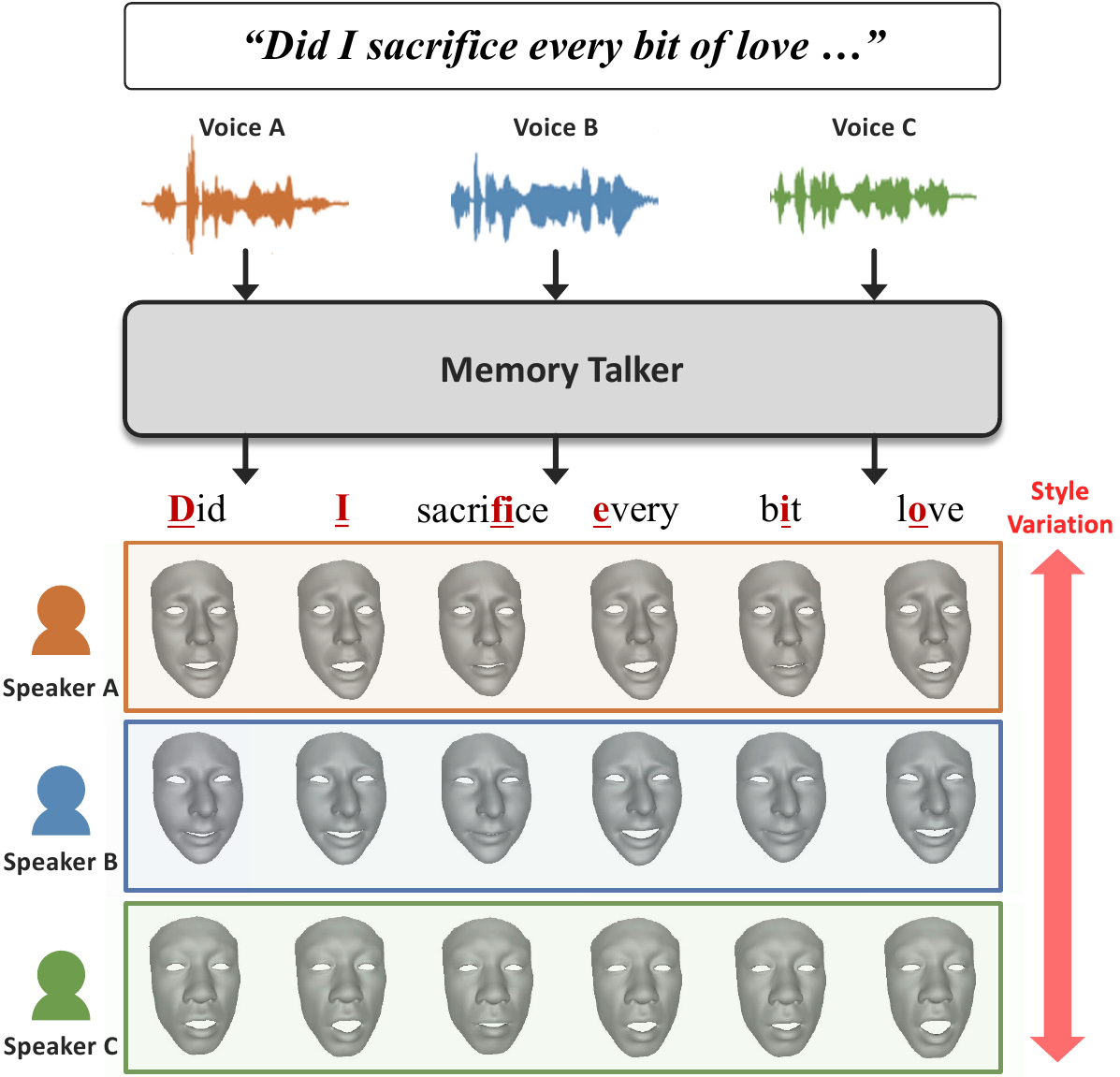}
\end{center}
% \vspace{-10pt}
\caption{The intuition of \textit{MemoryTalker} for personalized speech-driven 3D facial animation, \textit{Memorizing} and \textit{Animating}. 
\textbf{Memorizing}: Storing and retrieving facial motion.
\textbf{Animating}: Synthesizing the personalized 3D facial motion with the stylized motion memory. Our MemoryTalker can accurately produce the personalized 3D facial motion for different speakers using only audio input.}
\vspace{-10pt}
\label{fig:1}
\end{figure}
%##################################################################################################

%=========================== INTRODUCTION
\section{Introduction}
Speech-driven 3D facial animation is a challenging task that aims to synthesize realistic 3D facial motion synchronized with the given speech \cite{icip2024, facetalk2024, diffposetalk2024, geometry2021, kmtalk2024, scantalk2024, unitalker2024, zhuang2024learn2talk, chatziagapi2023avface, he2023speech4mesh, eungi2024enhancing, media2face24}.
This technique can be widely utilized in various immersive applications, including VR telepresence, character animation for film production and gaming. Recently, the advent of large-scale 3D facial animation datasets \cite{BIWI2010, VOCA2019, meshtalk2021, emotalk23, liu2024emage} and advancements in deep learning have significantly increased interest in this field.

The core challenge lies in developing algorithms that not only achieve precise speech-motion synchronization but also capture individual speaking styles - the subtle yet distinctive patterns in how different speakers articulate the same words. 

%------------------------------------ Fig_2
%##################################################################################################
\begin{figure*}[t!]
\begin{center}
\includegraphics[width=1.0\linewidth]{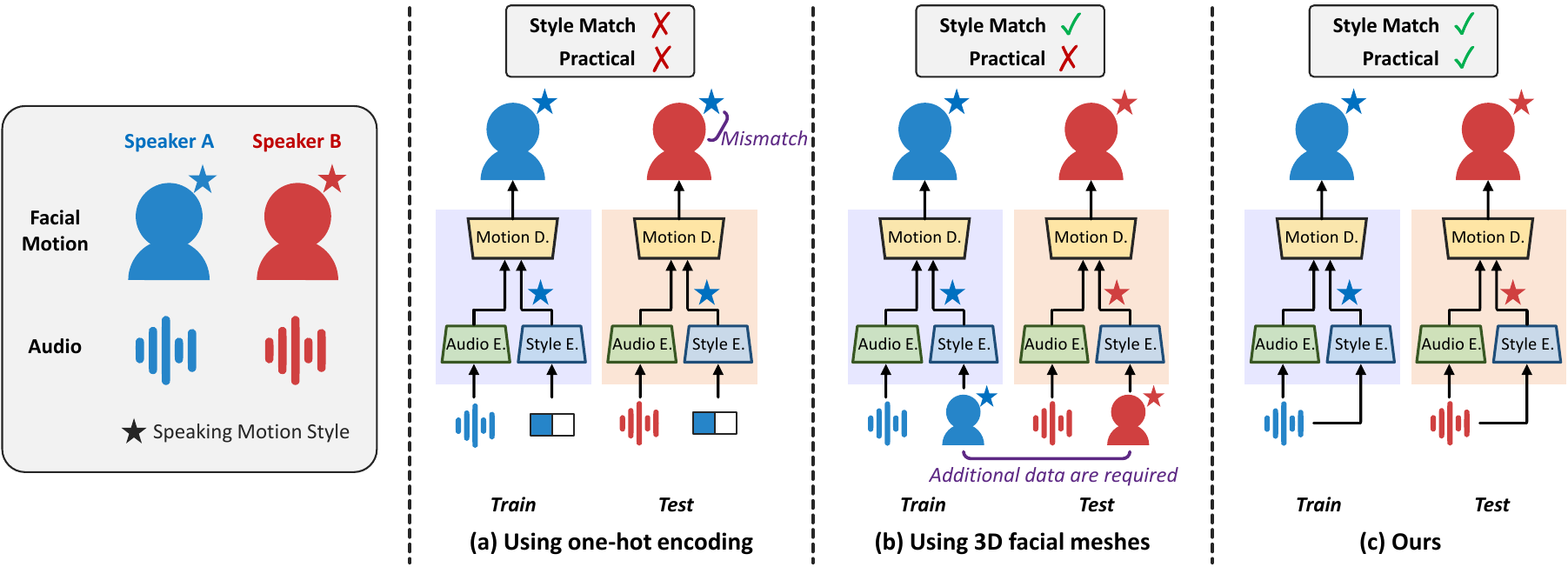}
% \vspace{-1.0cm}
\end{center}
    \caption{Explored approaches for personalized speech-driven 3D facial animation. 
    (a) In the one-hot encoding approaches, while the identities of training speakers can be encoded with a one-hot vector, it is impossible to match the unseen speaker's speaking style during inference. 
    (b) In the approaches of utilizing 3D facial mesh sequence, by providing an additional sequence of 3D facial mesh deformation during inference, they can produce the personalized 3D facial motion. However, it is not practical.
    (c) In our model, both during training and inference, personalized 3D facial animation is generated solely from the given audio input.}
\vspace{-10pt}
\label{fig:2}
\end{figure*}
%##################################################################################################

This work addresses \textit{personalization} issues in speech-driven 3D facial animation by considering the \textit{speaking style} of a speaker, including the amplitude of mouth opening and closing, the extent of pouting, etc. (see Fig. \ref{fig:1}).
To deal with that, most previous works \cite{VOCA2019, faceformer2022, codetalker2023, imitator2023} have employed one-hot encoding of the identity classes for different speakers in the training set.
However, these models necessarily require human identity classes even at inference time (Fig. \ref{fig:2} (a)). Furthermore, the inherent limitation of one-hot encoding makes it impossible for them to deal with unseen speakers at all.
The most recent studies \cite{CompositeandRegionalFacialMovements, probabilisticspeechdriven3dfacial23, mimic2024} have explored using a sequence of 3D facial mesh deformation to control the different speaking styles of each speaker.
Unlike one-hot encoding approaches, these methods can capture speaking styles of speakers by encoding the speaking style feature from their arbitrary facial motion sequences without class labels. Feeding 3D facial motion would ideally reflect the speaking style, but requiring such additional data at inference is not practical for real-world applications (Fig. \ref{fig:2} (b)).

To overcome these limitations, we propose a novel speech-driven 3D facial animation model named \textit{MemoryTalker} which can capture and predict speaking styles solely from audio input (see Fig. \ref{fig:2} (c)). It involves key-value multimodal memories that can effectively bridge different modalities. Our method consists of two training stages: $<$1-stage$>$ is for storing general motion into a motion memory (\textit{i.e.}, Memorizing), and $<$2-stage$>$ is for stylizing the motion memory to synthesize personalized animations (\textit{i.e.}, Animating).

In the first stage, we leverage a pre-trained ASR model to extract general motion feature representations with respect to the text aspect, ensuring consistent movements for a single phoneme. 
For example, when people say the word ``who”, the lips generally first come together and move forward to form the `W’ sound, then round to produce the `OO’ vowel sound. To explicitly store the facial motion features, we design a motion memory and access the stored motion features using the text representations with a key-value structure. This key-value memory allows the model to map one modality to another effectively through accessing the stored motion features with different modal features in completely separate feature spaces. It alleviates the domain gap from inconsistent distributions of different modalities. However, it is difficult to synthesize accurate facial motion only with text representations because there exist different speaking styles even for the same word ``who'' such as variations in the extent of pouting. 

To address this issue, in the second stage, our model is guided to generate personalized facial motions via audio-guided stylization. To this end, we personalize the trained motion memory based on audio signals. By distinguishing audio style features according to speaker types, we can achieve distinct style representations and refine the memory to synthesize the desired personalized motion effectively. Importantly, the proposed method does not require any prior knowledge (\textit{e.g.}, ID class, additional facial meshes) at inference time, which makes our model more practical for real-world applications.

\noindent Our contributions are summarized as follows: 

\begin{itemize}
    \item We propose \textit{MemoryTalker}, a novel framework for speech-driven 3D facial animation that can reflect speaking styles from audio input alone. To the best of our knowledge, this is the first work to address the personalization issue in this task without requiring additional prior information at inference time, which makes our approach practical.
    \item We introduce a new two-stage training strategy: (\textit{i}) memorizing general facial motions aligned with neutral text representations, and (\textit{ii}) animating personalized facial motions based on the stylized motion memory. This allows generating realistic animations that are accurate in terms of text content, while reflecting speaking styles effectively.
\end{itemize}
%===========================

%------------------------------------ Fig_3
%##################################################################################################
\begin{figure*}[t!]
\begin{center}
\includegraphics[width=0.80\linewidth]{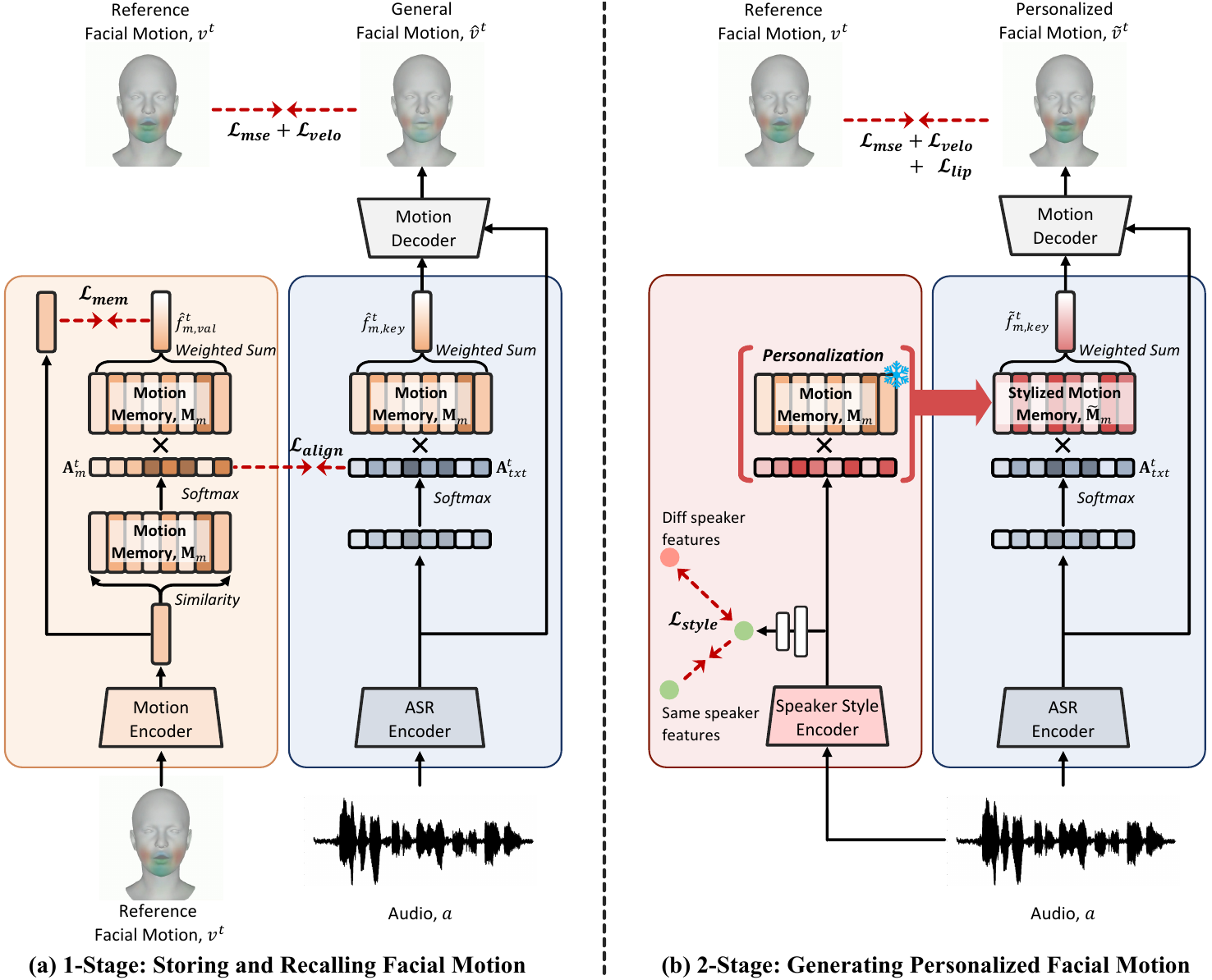}
\vspace{-0.5cm}
\end{center}
        \caption{Illustration of the proposed \textit{MemoryTalker} model for personalized 3D facial animation. (a) Learning to store facial motion feature in the facial motion memory and align motion features with text features. (b) Learning to disentangle unique speaking style of each speaker from audio and stylizing facial motion memory.}
\label{fig:method}
\vspace{-9pt}
\end{figure*}
%##################################################################################################

%=========================== RELATED WORK
\section{Related Work}
\subsection{Speech-Driven 3D Facial Animation} 
Earlier methods \cite{HMMbase2003, facial2006, jali2016} for 3D facial animation usually employ a predefined facial template consisting of 3DMM parameters. Recently, various studies have been conducted to consider not only the mapping of speech to 3D facial mesh movements but also to reflect identity-specific speaking styles. Existing works \cite{VOCA2019, meshtalk2021, faceformer2022, codetalker2023} have proposed a one-hot encoding of the identities of the training set. However, these models have a limitation in that they fail to predict new speaking styles during inference due to the use of the one-hot label slots for the subjects used in training.

To reflect subtle speaking styles better, Imitator \cite{imitator2023} introduces a two-stage style adaptation to synthesize speaker-independent movement in the first stage and capture speaker-dependent style in the second stage using reference 2D video. %4
To achieve careful control over speech-independent factors (e.g., speaking style), several methods have been proposed using reference 3D motion.
Wu et al. \cite{CompositeandRegionalFacialMovements} proposed an adaptive modulating module to consider both speech-independent composite and facial-dependent regional movements.
Mimic \cite{mimic2024} learned two disentangled latent spaces (content and style) for style-content disentanglement and 3D facial animation generation with identity-specific speaking styles. Finally, when generating 3D animation, they generate motion aligned to the audio in content space and apply a randomly provided speaking style in style space.
Yang et al. \cite{probabilisticspeechdriven3dfacial23} proposed a method to apply fine-detailed styles incrementally using reference 3D movements. Initially, coarse facial motion is learned from audio. Then, 3D facial motion with style is gradually learned from the provided reference 3D movement.
However, these methods require a reference 2D video or a sequence of 3D facial meshes as well as audio signals as inputs during inference, which are resource-consuming and impractical for real-world applications.
In contrast, we propose a novel approach that leverages speech-driven personalized memory networks to capture identity-specific speaking styles from speech signals alone.

\subsection{Memory Network} Memory networks can enhance inference capability by being read from and written with external long-term memory components \cite{weston2014memory, miller2016key}.
The key-value memory network architecture, enabling models to use keys to access relevant memories and retrieve corresponding values, has been widely adopted in various computer vision tasks including object tracking \cite{objecttrack18, objecttrack21}, few-shot learning \cite{fewshot18, fewshot20}, and anomaly detection \cite{anomalyD19, anomalyD21, 9496635, 10.1007/978-3-031-43412-9_12}.
In recent years, the memory networks have been used in multi-modal modeling \cite{predictiveL22, multimodalM21, multimodalM22, priyasad2021memory, 9541112}.
In \cite{yi2020audio, multimodalM22_2, tan2023emmn}, memory networks have been proposed to solve the alignment of associations between audio and visual information for 2D talking face generation.
Compared to the previous works, we introduce a novel two-stage training strategy that enables storing and retrieving general motion and then generating personalized facial animation with the motion memory stylized by the audio-driven speaking style feature. 
%============================

%=========================== PROPOSED METHOD ===========================
\section{Proposed Method}
Fig. \ref{fig:method} shows an overview of our personalized speech-driven 3D facial animation.
To synthesize realistic 3D facial animation, we propose a two-stage training strategy: 1) storing and recalling general facial motion and 2) generating personalized facial motion with the stylized motion memory.
In the first stage, our goal is to store facial motion features $f^t_m$ in the motion memory $\textbf{M}_m=\left\{s^i_m\right\}^n_{i=1}$ and recall general facial motion information from input audio $a^t$ at time $t$.
To achieve this, we encode the text representation $f^t_{txt}$ from $a^t$ and align it with $f^t_m$. This allows us to map the facial motion of various speakers for a single phoneme to the consistent text representation $f^t_{txt}$, thereby obtaining the general facial motion from the audio query $a^t$.
In the second stage, we synthesize the personalized facial motion reflecting the speakers' speaking styles from audio $a^{1:T}$.
To this end, we refine the pre-trained motion memory $\textbf{M}_m$ into the stylized motion memory $\tilde{\textbf{M}}_m$ while refining the text representation utilizing speaking style features $f_s$.
The speaking style feature learns distinct characteristics from speaker-specific audio by adopting a triplet loss.
Finally, the recalled personal motion features from the stylized motion memory $\tilde{\textbf{M}}_m$ are combined with the refined text representation and fed to the motion decoder to synthesize personalized 3D facial animation.

\subsection{Facial Motion Memory}
First, we design a motion memory $\textbf{M}_{m} \in \mathbb{R}^{n \times c}$ with $n$ slots and $c$ channels to store motion feature $f_m^t \in \mathbb{R}^c$.
The facial motion $v^t \in \mathbb{R}^3$ denotes the 3D movement of vertices over a neutral-face mesh template for the $t$-th frame \cite{codetalker2023}. 
To this end, $v^t$ is transformed into facial motion feature $f_m^t$ using a motion encoder $E_m$.
The encoded facial motion feature is used as a query to access our motion memory $\textbf{M}_{m}$.
When the facial motion feature $f_m^t$ is given as a query in the motion memory $\textbf{M}_m$, the attention weight $w_m^i$ can be obtained by calculating the similarity between the $f_m^t$ and each slot $s_m^i$ in $\textbf{M}_m$.
This addressing procedure can be formulated as
\begin{equation}
    w_m^i=\frac{\text{exp}(\kappa \cdot d(s_m^i, f_m^t))}{\sum_{j=1}^n{\text{exp}(\kappa \cdot d(s_m^j, f_m^t)}},
\end{equation}
where $d(\cdot,\cdot)$ indicates the cosine similarity function. $\kappa$ is a scaling factor. We omit the superscript $t$ here for simplicity.
%########comment 4-7################################
Let $\textbf{V}_m^t = \left\{w_m^1, w_m^2, \cdots, w_m^n \right\}$ denote the value address vector, which is a set of attention weights for the corresponding motion memory slots at time $t$.
The recalled motion feature $\hat{f}_{m,val}^t$ can be retrieved using $\textbf{V}_m^t$ and $\textbf{M}_{m}$ by a weighted sum for each memory slot as follows:
\begin{equation}
    \hat{f}_{m,val}^t = \sum_{i=1}^n w_{m}^{i} \cdot s_m^i.
\end{equation}

\noindent To explicitly embed motion features into the motion memory, we adopt the memory reconstruction loss between the recalled and reference motion features. 
Motion information is stored in $\textbf{M}_m$ by minimizing $\mathcal{L}_{mem}$.
\begin{equation}
    \mathcal{L}_{mem} = \sum_{t=1}^T\left\| f_m^t - \hat{f}_{m,val}^t \right\|_2^2.
\end{equation}

We search for motion features that are synchronized with the input audio after storing memory. 
At this time, to reduce the influence of individual speaking style variations in the audio (\textit{i.e.,} mapping various speaking styles to a single phoneme when the same pronouncing), we use the encoded text representation $f_{txt}^t$ from the encoder of automatic speech recognition (ASR) $E_{aud}$ as a query to access memory. 
This leads us to query common facial motion (i.e., common lip shapes for the same word regardless of the speaker).    %Common motion 작성 
The encoded text representation $f_{txt}^t$ can be written as
\begin{equation}
    f_{txt}^{1:T} = \psi_{\rightarrow n}(\text{Interp}(E_{aud}(a))) \in \mathbb{R}^{T\times n},
\end{equation}
% ##########comment 4-7##########
where $\psi_{\rightarrow n}(\cdot)$ is a single linear layer that projects to fit the number of slots.
% ###############################
$a$ represents the input audio signal. 
The pre-trained ASR encoder in HuBERT \cite{HuBERT2021} is used as $E_{aud}$ to map the source audio segment $a^t$ to the text representation $f_{txt}^t$. 
% ##########comment 4-7##########
$\text{Interp}(\cdot)$ is a linear interpolation function to synchronize the motion features and the text representations as video fps.

% ##############################
Let $\textbf{K}_{txt}^t=\left\{w_{txt}^1, w_{txt}^2, \cdots, w_{txt}^n \right\}$ denote the key address vector, which is a set of attention weights for the corresponding motion memory slots.
The key address vector $\textbf{K}_{txt}^t$ is obtained by applying a softmax function to the projected text representation $f_{txt}^t$. It can be written as
\begin{equation}
    \mathbf{K}_{txt}^t = \text{softmax}(f_{txt}^t) \in \mathbb{R}^n.
\end{equation}

The key address vector $\textbf{K}_{txt}^t$ derived from $f_{txt}^t$ is used to recall the general facial motion information across various speakers.
%when they say the same word.
Finally, the motion feature $\hat{f}_{m,key}^t \in \mathbb{R}^c$ is retrieved from motion memory using the text address aligned to the motion address for general facial motion synthesis.
\begin{equation}
    \hat{f}_{m,key}^t = \sum_{i=1}^n w_{txt}^{i} \cdot s_m^i.
\end{equation}

To recall the corresponding motion features from $\textbf{M}_m$ using key address vector $\textbf{K}_{txt}^t$, it is required to align the key address vector $\textbf{K}_{txt}^t$ with the value address vector $\textbf{V}_{m}^t$.
For this purpose, we employ KL divergence between them as an alignment loss $\mathcal{L}_{align}$, which can be defined as
\begin{equation}
    \mathcal{L}_{align} = \sum_{t=1}^T \textbf{K}_{txt}^t \log\frac{\textbf{K}_{txt}^t}{\textbf{V}_m^t}.
\end{equation}

\noindent To synthesize facial motion $v^t$, we employ a motion decoder $D_m(\cdot)$ based on a Transformer decoder structure as in \cite{faceformer2022}. 
The facial motion can be generated as follows:
\begin{equation}
    \hat{v}^t = D_m([f_{txt}^t; \hat{f}_{m,key}^t], f_{txt}^{t}).
\end{equation}

\noindent The motion decoder is trained to minimize the reconstruction loss between the synthesized 3D facial motion $\hat{v}^t$ and ground truth 3D facial motion $v^t$. It can be written as
\begin{equation}
    \mathcal{L}_{mse} = \sum_{t=1}^T \left\|v^t - \hat{v}^t \right\|^2.
\end{equation}

\noindent In addition, we introduce a velocity loss $\mathcal{L}_{vel}$ to address the issue of jittery output frames when using only reconstruction loss \cite{selftalk23}.
$\mathcal{L}_{vel}$ is defined as
\begin{equation}
    \mathcal{L}_{vel} = \sum_{t=0}^{T-1} \left\|(v^{t+1}-v^{t}) - (\hat{v}^{t+1}-\hat{v}^{t}) \right\|^2.
\end{equation}

\noindent The total loss at the first training stage is defined as
\begin{equation}
    \mathcal{L}_{\text{1-stage}} = \mathcal{L}_{mse} + \mathcal{L}_{vel} + \lambda_1(\mathcal{L}_{mem} + \mathcal{L}_{align}),
\end{equation}
where we set $\lambda_1$ to 0.01.

%------------------------------------ Table_1
%##################################################################################################
\begin{table*}[!t]
\centering
\small % 작은 글씨 크기 설정
\renewcommand{\arraystretch}{1.3} % 줄 간격 조정
\resizebox{\textwidth}{!}{ % 페이지 너비에 맞게 테이블 크기 조정
\begin{tabular}{lccccclcccccc}
\bottomrule
\multicolumn{1}{l}{}            & \multicolumn{5}{c}{VOCASET \cite{VOCA2019}}                                                                      &   & \multicolumn{5}{c}{BIWI \cite{BIWI2010}} \\ 
                                \cline{2-6}                                                                                                         \cline{8-12} 
Method                          & $\text{FVE}\downarrow$& $\text{LVE}\downarrow$& $\text{FID}\downarrow$& $\text{LDTW}\downarrow$& $\text{Lip-max}\downarrow$        &   & $\text{FVE}\downarrow$& $\text{LVE}\downarrow$& $\text{FID}\downarrow$   & $\text{LDTW}\downarrow$ & $\text{Lip-max}\downarrow$\\ 
                                & $(\times 10^{-6})$    & $(\times 10^{-5})$    & $(\times 10^{-1})$    & $(\times 10^{-5})$     & $(\times 10^{-4})$ &   & $(\times 10^{-4})$    & $(\times 10^{-4})$    & $(\times 10^{-1})$       & $(\times 10^{-4})$     & $(\times 10^{-3})$ \\ 
\bottomrule
FaceFormer \cite{faceformer2022}& 0.639                 & 0.413                 & 3.583                 & 0.507                  & 0.452            &   & 0.981                 & 0.207                 & 8.204                    & 0.114                  & 0.477 \\
CodeTalker \cite{codetalker2023}& 0.721                 & 0.498                 & 3.713                 & 0.554                  & 0.484            &   & 0.979                 & 0.211                 & 9.419                    & 0.120                  & 0.478 \\
SelfTalk \cite{selftalk23}      & 0.593                 & 0.382                 & 3.279                 & 0.475                  & 0.416            &   & 1.030                 & 0.222                 & 7.320                    & 0.118                  & 0.496 \\
Imitator \cite{imitator2023}    & 0.686                 & 0.456                 & 3.918                 & 0.554                  & 0.472            &   & -                     & -                     & -                        & -                      & -     \\
ScanTalk  \cite{scantalk2024}   & 0.609                 & 0.375                 & 3.623                 & 0.457                  & 0.420            &   & -                     & -                     & -                        & -                      & -     \\
UniTalker   \cite{unitalker2024}& 0.570                 & 0.382                 & 3.256                 & 0.507                  & 0.407            &   & 0.919                 & 0.196                 & 7.234                    & 0.109                  & 0.461 \\
\rowcolor{custom_gray}
\bf MemoryTalker         & \textbf{0.506}        & \textbf{0.293}        & \textbf{3.045}        & \textbf{0.418}         & \textbf{0.331}   &   & \textbf{0.901}        & \textbf{0.187}        & \textbf{7.202}           & \textbf{0.107}         & \textbf{0.398} \\
\toprule
\end{tabular}
}
\vspace{-0.3cm}
\caption{Quantitative evaluation for speech-driven 3D facial animation on VOCASET\cite{VOCA2019} and BIWI\cite{BIWI2010}.}
\vspace{-5pt}
\label{table:1}
\end{table*}
%##################################################################################################

\subsection{Stylized Motion Memory}
To synthesize the personalized 3D facial motion, the speaking style feature $f_s$ is encoded from the audio $a$.
Without loss of generality, we assume that the audio signals include inherently speaking styles such as volume, pitch, and speaking speed.
These speaking styles lead to differences in mouth opening size or the extent of pouting in 3D facial animation even when speakers say the same word or sentence.

To encode the speaking style feature $f_s$ from audio $a$, we use the mel-spectrogram, which provides a representation aligned with human auditory perception.
The speaking style feature $f_s \in \mathbb{R}^c$ can be defined as %f_speaker를 정의
\begin{equation}
    f_s=E_s(\phi^{a2m}(a)),
\end{equation}
where $E_s$ is a speaking style encoder and $\phi^{a2m}$ is a function that converts audio into a mel-spectrogram.

In particular, to encode more discriminating style features, we introduce a speaking style loss $\mathcal{L}_{style}$ using a triplet loss.
The speaking style loss is designed to learn a feature space where similar speaking styles are closer (i.e., minimize intra-class variation) to each other and different speaking styles are further apart (i.e., maximize inter-class variation).
\begin{equation}
    \mathcal{L}_{style} = \max \left( \| f_s - f_s^p \|_2^2 - \| f_s - f_s^n \|_2^2 + l, 0 \right),
\end{equation}
% #################comment 4-7####################
where $f_s^p$ and $f_s^n$ are the speaking styles of positive samples (i.e., same speaker) and negative samples (i.e., different speakers), respectively. They are obtained through the same process as $f_s$. 
$l$ indicates a margin.
% ##################################

With the speaking style feature $f_s$, the pre-trained motion memory $\textbf{M}_m$ can be updated to the stylized motion memory $\tilde{\textbf{M}}_m = \{ \tilde{s}_m^i \}_{i=1}^n \in \mathbb{R}^{n \times c}$, which can be written as
\begin{equation}
    \tilde{\textbf{M}}_m = \{ \tilde{w}_s^i \cdot s_m^i \}_{i=1}^n,
\end{equation}
where $\tilde{\textbf{M}}_m$ is obtained by multiplying the style weight $\tilde{w}_s^i$ with the $i$-th memory slot in $\textbf{M}_{m}$.

To update each memory slot, we design the style weight $\tilde{w}_s = \left\{\tilde{w}^i_s\right\}^n_{i=1} \in \mathbb{R}^{n}$ as follows:
\begin{equation}
\tilde{w}_{s} = \text{sigmoid}(\psi'_{\rightarrow n}(f_s)) \cdot \psi_{\rightarrow 1}(f_s)
\end{equation}
where \(\psi'_{\rightarrow n}\) and \(\psi_{\rightarrow 1}\) are single linear layers that project to fit a \(n\)-dimensional vector and a scalar value, respectively. The values obtained from \(\psi'_{\rightarrow n}(f_s)\) are processed with the sigmoid function to score each slot, and then scaled by the result of \(\psi_{\rightarrow 1}(f_s)\).
By applying the $\tilde{w}_s$ to the motion memory $\textbf{M}_m$, the stylized motion memory $\tilde{\textbf{M}}_m$ can reflect different speaking styles for each speaker via audio only.    %6
%w_s가 memory slot마다 speaker 고유의 가중치를 의미한다는 말이 구체적으로 생략되어있어 reviewer가 질문을 하신 것 같습니다.

Finally, the personalized motion $\tilde{v}^t \in \mathbb{R}^3$ can be reconstructed by incorporating the personalized motion feature $\tilde{f}_{m,key}$ recalled in $\tilde{\textbf{M}}_m$. 
\begin{equation}
    \tilde{v}^t = D_m([f_{txt}^t; \tilde{f}_{m,key}^t], f_{txt}^t),
\end{equation}
where $\tilde{f}_{m,key}^t = \sum_{i=1}^n w_{txt}^{i} \cdot \tilde{s}_m^i$, which is computed in the same way as the weighted sum used to calculate $f_{m,key}^t$.

As a result, by explicitly embedding the speaking style features into the motion memory and recalling it, we can produce more realistic and personalized 3D facial animation.

In the second stage, to train our model, we additionally employ a lip vertex loss $\mathcal{L}_{lip}$, which is a mean square error for the lip region (i.e., lower-face vertices) to focus on carefully synthesizing lip movements according to various speaking styles.
The total loss for the second stage is defined as
\begin{equation}
    \mathcal{L}_{\text{2-stage}} = \mathcal{L}_{mse} + \mathcal{L}_{vel}  + \lambda_2(\mathcal{L}_{lip} + \mathcal{L}_{style}),
\end{equation}
where we set $\lambda_2$ to 0.01.
%============================

%=========================== EXPERIMENTS
\section{Experiments}

%------------------------------------ figure_4
%##################################################################################################
\begin{figure*}[t!]
\centering
\includegraphics[width=0.95\linewidth]{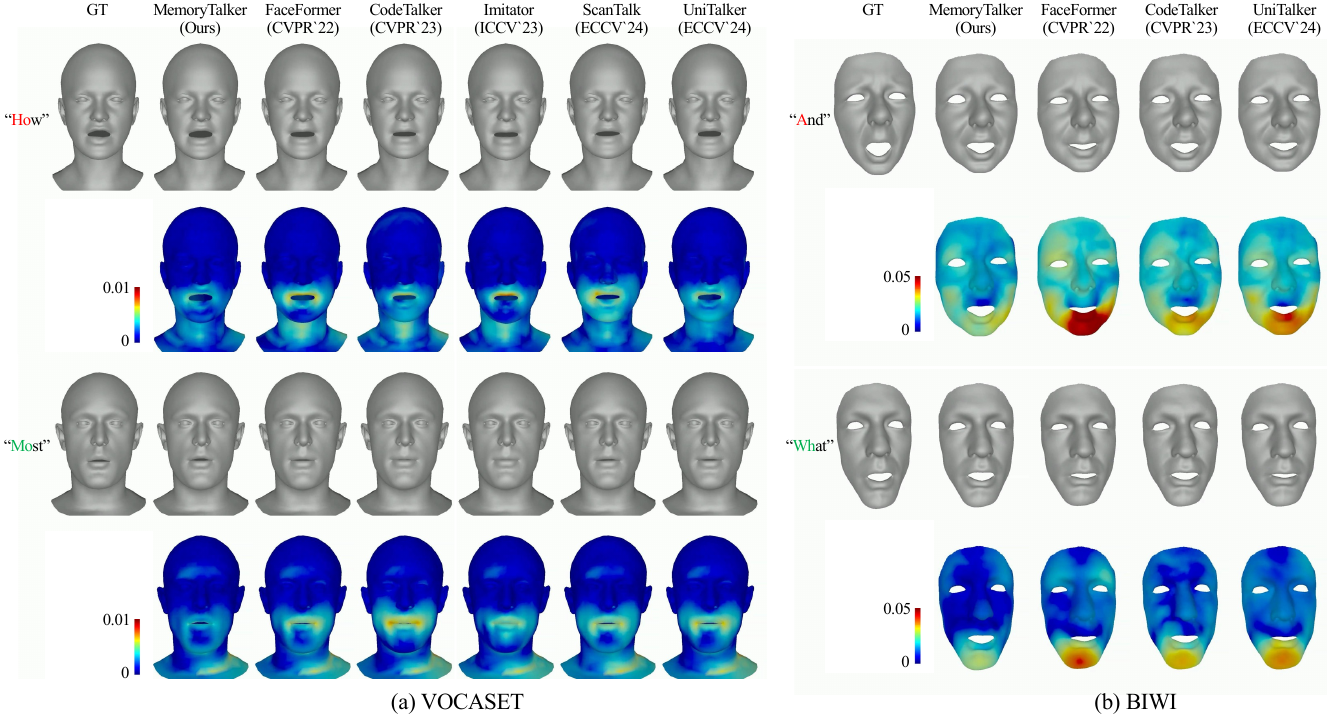}
\vspace{-10pt}
\caption{Visual comparisons with state-of-the art methods on (a) VOCASET and (b) BIWI. Note that the second and fourth rows represent the visualization of per-vertex errors. Note that we displayed the best one-hot encoding results \cite{faceformer2022,codetalker2023,imitator2023} for the ID with the lowest error.}
\vspace{-10pt}
\label{fig:4}
\end{figure*}
%##################################################################################################

% \subsection{Experiments Setup} 
\subsection{Datasets}
We train and evaluate our model on VOCASET \cite{VOCA2019} and BIWI \cite{BIWI2010}, which are widely used datasets for 3D facial animation. Both datasets contain audio-3D facial scan pairs that demonstrate English speech pronunciation.
VOCASET \cite{VOCA2019} consists of 255 unique sentences, some shared across speakers, and contains 480 facial motion sequences from 12 subjects, captured at 60 fps for approximately 4 seconds per sequence. Each 3D face mesh is registered to the FLAME \cite{FLAME2017} topology with 5,023 vertices.
BIWI \cite{BIWI2010} includes 40 unique sentences shared across all speakers.
It consists of two parts: one with emotions and one without. There are 40 sentences uttered by 14 subjects. Each recording is repeated twice in neutral or emotional situations, capturing a dynamic 3D facial scan at 25 fps. The registered topology exhibits 23,370 vertices, with the average sequence length of 4.67s.

\subsection{Implementation Details}
Our experiments were conducted on an NVIDIA A6000 GPU. A single linear layer is used as motion encoder $E_m$ as in \cite{faceformer2022, codetalker2023}. 
The structure in \cite{interspeech22} is employed as our style encoder $E_s$. To encode the text representation from the audio, we utilize the pre-trained ASR model \cite{HuBERT2021}.
The projection layer $\psi_{\rightarrow n}$ in ASR consists of a single linear layer to project audio features onto the text logit and is fine-tuned during 1-stage training.
In the first stage, we train our model (except for $E_s$)  for 100 epochs with a learning rate of 0.0001. 
In the second stage, we freeze all layers of the first stage and train only the speaking style encoder $E_s$ with a learning rate of 0.00005 for 100 epochs.
Quantitative evaluations were performed on VOCA-Test and BIWI-Test-B datasets by averaging the identities of all speakers.

\subsection{Evaluation Metrics} We adopt five quantitative evaluation metrics to evaluate the results: Face Vertex Error (FVE) and Lip Vertex Error (LVE) are the differences between the reference vertices and the generated vertices for the entire face and lip region, respectively. Lip Dynamic Time Warping (LDTW) is used to compute the temporal similarity of the lip region using DTW \cite{dtw07} as in \cite{imitator2023}. Fréchet Inception Distance (FID) score \cite{fid_ori17} is used to evaluate the quality of the images rendered from the vertices \cite{fid23}. In addition, we use Lip-max that averages the highest error among lip regions \cite{meshtalk2021, imitator2023}.

%------------------------------------ Table_2
%##################################################################################################

\begin{table}[!t]
\centering

\raggedright
\renewcommand{\arraystretch}{1.3} % 줄 간격 조정
% \begin{minipage}{0.5\textwidth} % 테이블 크기를 페이지 너비의 60%로 조정
% \renewcommand{\tabcolsep}{0.8mm}
\resizebox{\linewidth}{!}{%
\begin{tabular}{lcccc}
\bottomrule
\multirow{2}{*}{Method}             & $\text{FVE}\downarrow$        & $\text{LVE}\downarrow$        & $\text{FID}\downarrow$    \\
                                    & $(\times 10^{-6})$            & $(\times 10^{-5})$            & $(\times 10^{-1})$        \\ 
\bottomrule
FaceFormer \cite{faceformer2022}                          & $0.639^ {\pm 0.036}$             & $0.413^ {\pm 0.058}$             & $3.583^{\pm 0.358}$         \\
CodeTalker \cite{codetalker2023}                          & $0.721^ {\pm 0.056}$             & $0.498^ {\pm 0.037}$             & $3.713^{\pm 0.373}$         \\
Imitator \cite{imitator2023}                           & $0.686^{\pm 0.069}$             & $0.456^ {\pm 0.067}$             & $3.918^{\pm 0.536}$         \\
\rowcolor{custom_gray}
\bf MemoryTalker                 & \textbf{0.506}                & \textbf{0.293}                & \textbf{3.045}            \\
\toprule
\end{tabular}
}
\vspace{-0.3cm}
\caption{Error variability according to the used one-hot identity about a given audio on VOCASET \cite{VOCA2019}.}
% \vspace{-5pt}
\label{table:2}
% \end{minipage}
\end{table}

%##################################################################################################

%------------------------------------ Table_efficiency
%##################################################################################################
\begin{table}[!t]
\centering
\renewcommand{\arraystretch}{1.3}
\renewcommand{\tabcolsep}{5mm}
\resizebox{\linewidth}{!}{
\begin{tabular}{lcc}
\bottomrule
Method & Inference Time & Parameter \# \\
\bottomrule
FaceFormer \cite{faceformer2022} & 38.1 ms & {92} M\\
CodeTalker \cite{codetalker2023} & 297.6 ms & 315 M\\
SelfTalk \cite{selftalk23}       & 10.1 ms & 450 M \\
UniTalker \cite{unitalker2024}   & 9.7 ms & 313 M \\
\rowcolor{custom_gray} \textbf{MemoryTalker} & {7.8} ms & 94 M\\
\toprule
\end{tabular}
}
\vspace{-0.3cm}
\caption{Efficiency comparison of models based on inference time and number of parameters on VOCASET\cite{VOCA2019}.}
\vspace{-5pt}
\label{table:efficiency}
\end{table}
%##################################################################################################

\subsection{Quantitative Results}
Tab. \ref{table:1} illustrates that our MemoryTalker outperforms state-of-the arts on both datasets. 
Our model achieves the lowest prediction errors.
The lowest LDTW of our MemoryTalker means the highest temporal similarity for the lip region.
Tab. \ref{table:2} illustrates the problem of one-hot encoding approaches. They employ various identities in the training set for single audio during inference. Thus, they cannot reflect the speaking style of an unseen speaker during inference.
As a result, the rendering quality for the same speaker can vary depending on the identity selected from the training set. 
On the other hand, our MemoryTalker can reflect the audio-driven speaking style feature, matching the speaker. 

To evaluate the computational efficiency, as seen in Tab. \ref{table:efficiency}, we measure the number of learnable parameters and inference time with a 1-second audio sample on VOCASET \cite{VOCA2019}. Compared to FaceFormer \cite{faceformer2022}, CodeTalker \cite{codetalker2023}, SelfTalk \cite{selftalk23}, and UniTalker \cite{unitalker2024}, our MemoryTalker achieves efficient 3D facial animations with about 120 fps.

% ####################################################################################

%------------------------------------ Figure_5
%##################################################################################################
\begin{figure}[t!]
\centering
\includegraphics[width=1.0\linewidth]{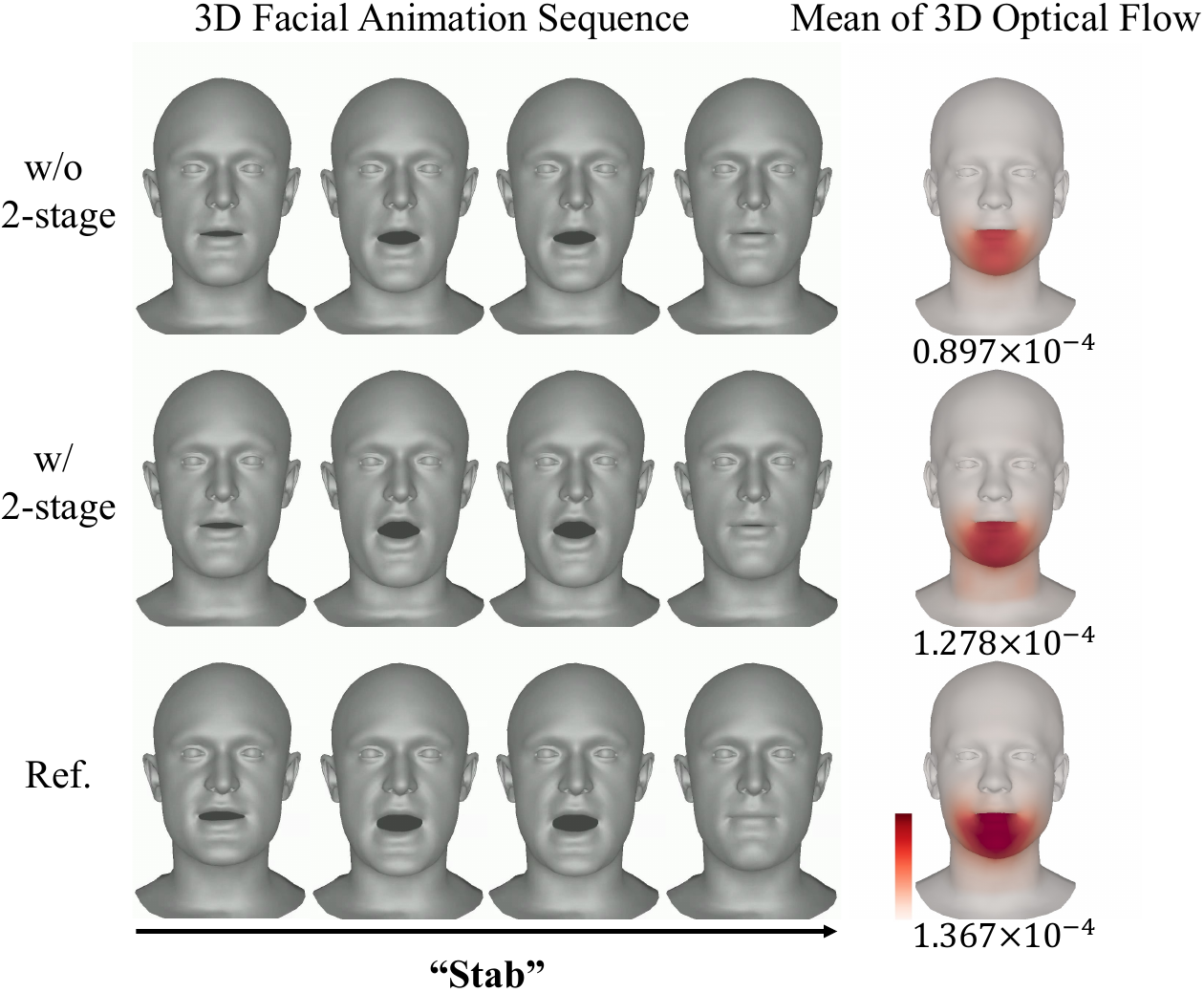}
\caption{Qualitative results from ``w/o 2-stage'' and ``w/ 2-stage''. The last column represents the mean of 3D optical flow while pronouncing ``\textbf{Stab}''. The darker the red, the higher the motion.}
\label{fig:5}
% \vspace{-5pt}
\end{figure}
%##################################################################################################

%------------------------------------ Figure_6
%##################################################################################################
\begin{figure}[t!]
\centering
\includegraphics[width=1\linewidth]{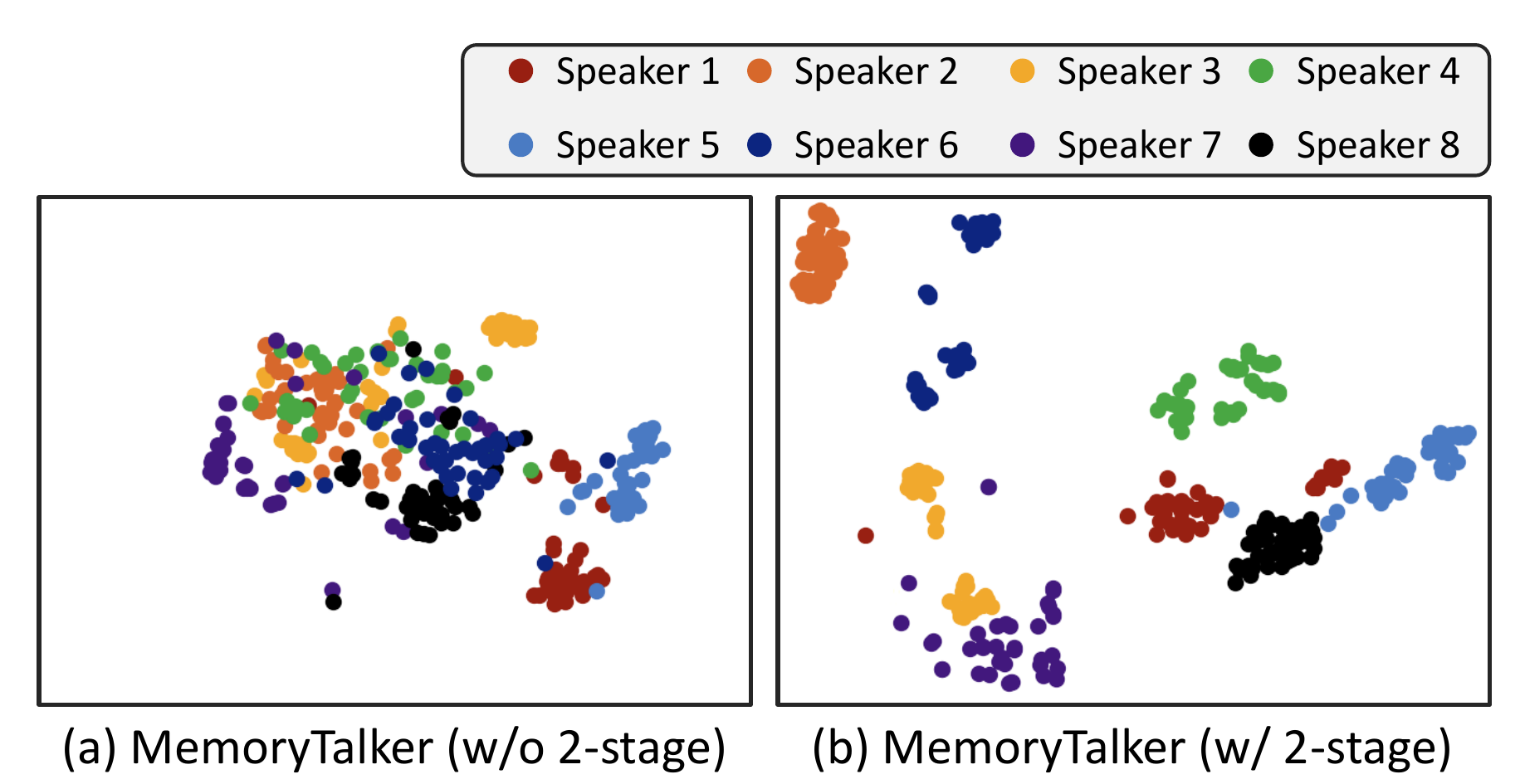}
\caption{The t-SNE visualization of the recalled motion features across different speakers at each stage.}%(1-4)
\vspace{-5pt}
\label{fig:6}
\end{figure}
%##################################################################################################

\subsection{Qualitative Results}
\subsubsection{Visual Comparisons}
Fig. \ref{fig:4} shows the visual results of our MemoryTalker, compared to existing methods.
When pronouncing ``\textcolor{red}{Ho}w'' and ``\textcolor{red}{An}d'' with a mouth open, our results provide more accurate mouth shapes.
For pronunciations with mouth protrusion like ``\textcolor[rgb]{0.31, 0.69, 0.37}{Mo}st'', ``\textcolor[rgb]{0.31, 0.69, 0.37}{Wh}at'', our results show similar mouth shapes to the reference (i.e., ground-truth).
Although recent models generate some extent of mouth opening and closing according to the pronunciation, they lead to larger prediction errors around lip regions (lower-face vertices), compared to our MemoryTalker.

\subsubsection{Adaptation of Speaking Style}
To verify the effectiveness of our stylized motion memory at the 2-stage, we evaluate the dynamics of facial vertices sequence using 3D optical flows.
Fig. \ref{fig:5} shows an example of visual results and the mean of optical flows from 1-stage and 2-stage when pronouncing ``\textbf{Stab}''.
As shown in Fig. \ref{fig:5}, the model trained at both stages delivers more accurate lip movements than the one trained at 1-stage only (without training the model at 2-stage).

To explore the effectiveness of the stylized motion memory in the latent space, we visualize the recalled features before decoding each subject's facial motion at each stage using t-SNE \cite{t-SNE08}.
In Fig. \ref{fig:6}, each point represents facial motion features for each sentence, and different colors indicate different speakers.
Fig. \ref{fig:6} (a) shows that the distribution of each speaker is not separated because general facial motion is retrieved at 1-stage.
In Fig. \ref{fig:6} (b), the distribution for each speaker is well-clustered, indicating that the speaking style of each speaker is captured at 2-stage.

%------------------------------------ Table_4
%##################################################################################################

\begin{table}[!t]
\renewcommand{\arraystretch}{1.3}
\renewcommand{\tabcolsep}{5mm}
\centering
\resizebox{\linewidth}{!}{
\begin{tabular}{cccc}
\bottomrule
\multirow{2}{*}{\makecell{Proposed \\ 1-stage training}} & \multirow{2}{*}{\makecell{Proposed \\ 2-stage training}} & $\text{FVE}\downarrow$ & $\text{LVE}\downarrow$ \\
                                                    &                          & $(\times 10^{-6})$     & $(\times 10^{-5})$ \\
\bottomrule
 \textcolor{darkred}{\ding{55}} & \textcolor{darkred}{\ding{55}}     & 0.638 & 0.460 \\
 \textcolor{darkgreen}{\ding{51}} & \textcolor{darkred}{\ding{55}}   & 0.531 & 0.313 \\
\rowcolor{custom_gray}
 \textcolor{darkgreen}{\ding{51}} & \textcolor{darkgreen}{\ding{51}} & \bf 0.506 & \bf 0.293 \\ 
\toprule
\end{tabular}
}
\caption{Ablation study for various modules on VOCASET \cite{VOCA2019}.}
\vspace{-10pt}
\label{table:3}
\end{table}

%##################################################################################################

\subsection{Ablation Study}
\subsubsection{Memory and Style Encoder}
To demonstrate the effectiveness of two-stage training, we evaluate the performance at each stage.
In Tab. \ref{table:3}, the baseline (w/o 1-stage and 2-stage) is a model without our motion memory structure and is trained by adopting common reconstruction loss $\mathcal{L}_{mse}$ and velocity loss $\mathcal{L}_{vel}$. 
As seen in Tab. \ref{table:3}, by explicitly storing and recalling the facial motion in 1-stage, FVE and LVE are reduced.
It means that the facial motion recalled from the facial motion memory can support the facial synthesis results from text representation. 
In addition, by incorporating the personalized facial motion through the stylized motion memory in 2-stage, our model achieves the best performance.

\subsubsection{Constraints}
We investigate the impact of removing different constraints which are $\mathcal{L}_{lip}$ and $\mathcal{L}_{style}$ (see Tab. \ref{table:4}).
When removing the $\mathcal{L}_{lip}$, performance is slightly worse than those with $\mathcal{L}_{lip}$.
This is because the lip vertex loss is only relevant to the lower-face vertices.
When removing the triplet loss for $\mathcal{L}_{style}$, we can observe that training becomes unstable and performance drops significantly. It highlights the importance of effectively learning personal speaking styles.

%------------------------------------ Table_5
%##################################################################################################

\begin{table}[!t]
\renewcommand{\arraystretch}{1.3}
\centering
\raggedright
\resizebox{\linewidth}{!}{
{\scriptsize
\begin{tabular}{lcccc}
\bottomrule
\multirow{2}{*}{Loss}                           & $\text{FVE}\downarrow$ & $\text{LVE}\downarrow$ & $\text{FID}\downarrow$ & $\text{LDTW}\downarrow$  \\
                                                & $(\times 10^{-6})$     & $(\times 10^{-5})$     & $(\times 10^{-1})$     & $(\times 10^{-5})$       \\ 
\bottomrule

\multicolumn{1}{l}{w/o $\mathcal{L}_{lip}$}     & 0.514                  & 0.297                  & 3.057                  & 0.435                    \\
\multicolumn{1}{l}{w/o $\mathcal{L}_{style}$}   & 0.513                  & 0.295                  & 3.058                  & 0.431                    \\ 
\rowcolor{custom_gray}
\textbf{Full}                                   & \bf 0.506              & \bf 0.293              & \bf {3.045}            & \bf 0.418                \\
\toprule
\end{tabular}
}
}
\caption{Ablation study for our components on VOCASET \cite{VOCA2019}.}
\vspace{-7pt}
\label{table:4}
\end{table}
%##################################################################################################

%------------------------------------ Table_6
%##################################################################################################

\begin{table}[!t]
\renewcommand{\arraystretch}{1.3}
\renewcommand{\tabcolsep}{0.9mm}
\centering
\resizebox{\linewidth}{!}{% % Resize to 50% of the text width
\begin{tabular}{lccc}
\toprule
{Competitors}                           & Lip Sync(\%)  & Realism(\%)   & Speaking Style(\%)    \\
\bottomrule
vs. FaceFormer \cite{faceformer2022}    & 83.9      & 85.5       & 80.6             \\
vs. CodeTalker \cite{codetalker2023}    & 85.5      & 83.9       & 71.8             \\
% vs. SelfTalk   \cite{selftalk23}        & 70.2      & 71.0       & 81.5             \\
vs. Imitator   \cite{imitator2023}      & 87.1      & 87.1       & 78.2             \\
vs. ScanTalk   \cite{scantalk2024}      & 94.4      & 91.1       & 92.8             \\
vs. UniTalker   \cite{unitalker2024}    & 79.8      & 80.6       & 86.3             \\
\bottomrule
\end{tabular}
}
\caption{User study: our method vs. competitors on  VOCASET \cite{VOCA2019}.}
\label{table:5}
\vspace{-7pt}
\end{table}
%##################################################################################################

\subsection{User Study}
We conduct A/B tests on VOCASET to evaluate perceptual lip-sync, realism, and speaking style of synthesized 3D facial animations.
A total of 33 subjects participated in the user study.
To evaluate speaking style, lip-sync, and realism, we compare our MemoryTalker with 5 other methods (FaceFormer, CodeTalker, Imitator, ScanTalk, UniTalker) on 50 samples.
To measure lip-sync and realism, we present each pair of identical sentences to the subject and the subject chooses the better one.
To evaluate speaking style, we present participants with samples from our model, a comparison model, and the ground truth (reference). Participants are then asked to choose which of the two model-generated samples appears to have a speaking style most similar to the ground truth, considering factors such as the amplitude of mouth opening and closing, the degree of pouting, and other relevant aspects.
As shown in Table \ref{table:5}, our results are generally favorable over other state-of-the-art methods across a variety of perceptual factors.
%============================

%=========================== CONCLUSION
\section{Conclusion}
In this study, we demonstrate the ability of motion memory networks to bridge different modalities, which are 3D facial motion and speech, without additional information for personalized speech-driven facial animations. 
To effectively train our model, we introduce a two-stage training strategy: 1) storing and retrieving general motion and 2) performing the personalized 3D facial motion synthesis with the stylized motion memory. 
Extensive experiments highlight the superior performance of our MemoryTalker compared to the recent speech-driven 3D facial animation models.
In particular, our MemoryTalker achieves more favorable results not only in quantitative evaluations but also in the user study from a perceptual perspective.
We hope this work can pave the way in real-world VR and metaverse applications.
% We hope this work can pave the way for practical personalized speech-driven 3D facial animation in real-world VR and metaverse applications.
%============================

\section{Acknowledgments}
This research was supported by Culture, Sports and Tourism R\&D Program through the Korea Creative Content Agency grant funded by the Ministry of Culture, Sports and Tourism in 2023 (Project Name: Acquisition of 3D precise information of microstructure and development of authoring technology for ultra-high prediction cultural restoration, Project Number: RS-2023-00227749) and the National Research Foundation of Korea(NRF) grant funded by the Korea government(MSIT) (RS-2025-00563942).

% \newpage

% {
    % \small
    \bibliographystyle{ieeenat_fullname}
    \bibliography{main}
% }

\clearpage
\setcounter{page}{1}
\maketitlesupplementary
\setcounter{section}{0} % 섹션 번호 초기화
\renewcommand{\thesection}{\Alph{section}} % Supplementary에서는 A, B, C로 변경
\setcounter{figure}{0} % 그림 번호 초기화
\renewcommand{\thefigure}{S\arabic{figure}}
\setcounter{table}{0} % 테이블 번호 초기화
\renewcommand{\thetable}{S\arabic{table}}

\definecolor{custom_gray}{gray}{.92}

%=========================== ABSTRACT===========================
% \begin{abstract}
\section{Overview}

This supplementary document provides additional experiments and explanations to complement the main manuscript. We present a comprehensive performance comparison focusing on the effects of different memory slot sizes in our model. Additionally, we offer quantitative results comparing end-to-end training versus our proposed 2-stage training strategy. The document also includes extensive qualitative results, featuring detailed frame-by-frame comparisons between our method and existing approaches for different speakers under various conditions. Furthermore, we provide an in-depth feature analysis, visualizing how our method effectively captures speaking styles from audio input effectively. Lastly, we detail the implementation specifics of our MemoryTalker architecture and describe the methodology of our user study for a thorough evaluation against other methods.
%=================================================================================

%=========================== Performance Comparison
\section{Quantitative Results}
\subsection{Effects of Memory Slot Size}
\begin{figure}[b]
\includegraphics[width=1.0\linewidth]{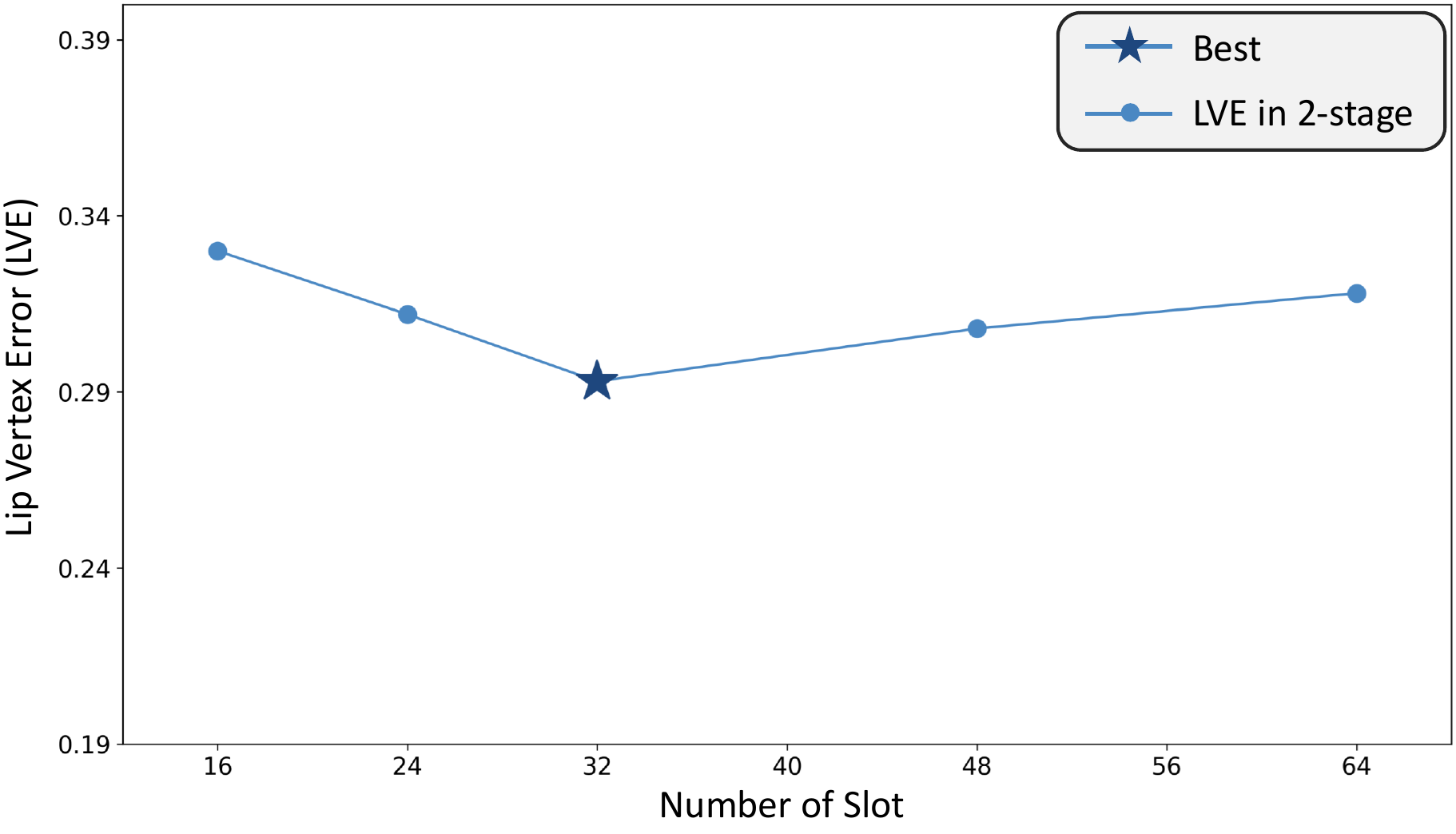}
% \vspace{-0.2cm}
\caption{Effects of the memory slot size on LVE for facial animation generation.}
% \vspace{-10pt}
\label{supp_fig:1}
\end{figure}

We conduct experiments varying the number of memory slots. Figure \ref{supp_fig:1} shows Lip Vertex Errors (LVE) across different slot size configurations. We validate the performance while varying the memory slot size to (16, 24, 32, 48, 64). The optimal performance is achieved with 32 slots (indicated by star marker in Figure \ref{supp_fig:1}). Note that we leverage memory slot size 32 for our experiments in the main manuscript.

%------------------------------------ Figure_1
%##################################################################################################
%##################################################################################################

\subsection{End-to-End Learning vs. Two-stage Training Strategy}
We propose a novel two-stage training strategy that first induces general motion synthesis, followed by speaking style adaptation. Table \ref{supp_table:1} shows performance comparisons between end-to-end learning and our two-stage training strategy. As seen in the Table \ref{supp_table:1}, our training strategy consistently outperforms the end-to-end learning across all evaluation metrics, including FVE, LVE, and FID. These results validate that our strategy more effectively captures and reproduces personalized speaking styles while maintaining motion accuracy.

%------------------------------------ Table_S1
%##################################################################################################
\begin{table}[t!]
\centering
\renewcommand{\arraystretch}{1.4} % 줄 간격 조정
\renewcommand{\tabcolsep}{4mm} % 열 간격 조정
\resizebox{\linewidth}{!}{
\begin{tabular}{ccccc} % 테이블 전체 너비를 조정
\bottomrule
\multirow{2}{*}{Training Strategy}              & $\text{FVE}\downarrow$        & $\text{LVE}\downarrow$        & $\text{FID}\downarrow$    \\
                                                & $(\times 10^{-6})$            & $(\times 10^{-5})$            & $(\times 10^{-1})$        \\ 
\bottomrule
End-to-End                                      & 0.510                         & 0.303                         & 3.142                     \\
% \rowcolor{LightBlue}
\rowcolor{custom_gray}
\bf Two-Stage (Ours)                              & \textbf{0.506}                & \textbf{0.293}                & \textbf{3.045}            \\
\toprule
\end{tabular}
}
\caption{Performance comparison between end-to-end learning and two-stage training strategy.}

\label{supp_table:1}
\end{table}
%##################################################################################################
%===========================

%------------------------------------ tableS2
%##################################################################################################
\begin{table}[!t]
\centering
\renewcommand{\arraystretch}{1.4} % 줄 간격 조정
\renewcommand{\tabcolsep}{1.5mm} % 열 간격 조정
\resizebox{\linewidth}{!}{
% 수정: 열 지정자 변경 lccccc -> lcccc
\begin{tabular}{ccccc} % 표 좌우의 추가 공간 제거
\toprule
% & \multicolumn{4}{c}{VOCASET \cite{VOCA2019}} \\
% \cline{2-5}
\multirow{2}{*}{Method} & $\text{FVE}\downarrow$ & $\text{LVE}\downarrow$       & $\text{LDTW}\downarrow$         & $\text{Lip-max}\downarrow$ \\
                        & $(\times 10^{-6})$     & $(\times 10^{-5})$           & $(\times 10^{-5})$              & $(\times 10^{-4})$                  \\ 

\midrule
Mesh-driven \cite{CompositeandRegionalFacialMovements} & 0.673 & 0.400 & 0.521 & 0.431 \\
\rowcolor{gray!15} % 행 배경색 (xcolor 패키지 필요)
\textbf{Audio-driven (Ours}) & \textbf{0.506} & \textbf{0.293} & \textbf{0.418} & \textbf{0.331} \\
\bottomrule
\end{tabular}
}
% \vspace{-0.3cm} % 필요에 따라 주석 해제 또는 값 조절
% 권장: 표에 대한 간단한 설명 추가
\caption{Quantitative results about mesh-based method.}
\label{tab:vocaset_comparison_rebuttal_no_fid} % 레이블 수정 (선택 사항)
\end{table}
%##################################################################################################

\subsection{Comparison with Mesh-based Methods}
As presented in Table~\ref{tab:vocaset_comparison_rebuttal_no_fid}, we conduct a quantitative comparison with an additional mesh-based method~\cite{CompositeandRegionalFacialMovements}. 
The results clearly indicate that our audio-driven approach achieves superior performance across all evaluation metrics. 
It is worth noting that Mimic~\cite{mimic2024}, another mesh-based method, could not be directly compared as it utilizes different datasets for training and evaluation.

%------------------------------------ tableS3
%##################################################################################################
\begin{table}[!t]
\centering
\renewcommand{\arraystretch}{1.4} % 줄 간격 조정
\renewcommand{\tabcolsep}{1.5mm} % 열 간격 조정
\resizebox{\linewidth}{!}{ % 현재 컬럼 너비에 맞게 테이블 크기 조정 (한 페이지 rebuttal에 적합)
% 수정: 열 지정자 변경 lccccc -> lcccc
\begin{tabular}{ccccc} % 표 좌우의 추가 공간 제거
\toprule
% & \multicolumn{4}{c}{VOCASET \cite{VOCA2019}} \\
% \cline{2-5}
\multirow{2}{*}{Method} & FVE$\downarrow$    & LVE$\downarrow$       & LDTW$\downarrow$           & Lip-max$\downarrow$    \\
                        & $(\times 10^{-1})$        & $(\times 10^{-3})$           & $(\times 10^{-5})$                & $(\times 10^{-1})$            \\ 
\midrule
CodeTalker \cite{codetalker2023} & 0.122 & 0.453 & 0.637 & 0.356 \\
UniTalker \cite{unitalker2024} & 0.101 & 0.283 & 0.569 & 0.305 \\
\rowcolor{gray!15} % 행 배경색 (xcolor 패키지 필요)
\textbf{MemoryTalker (Ours)} & \textbf{0.094} & \textbf{0.266} & \textbf{0.557} & \textbf{0.285} \\
\bottomrule
\end{tabular}
}
% \vspace{-0.3cm} % 필요에 따라 주석 해제 또는 값 조절
\caption{Quantitative results about seen identities.} % 캡션에 Params 언급 추가
\label{tab:biwi_seen_id} % 레이블 수정 (선택 사항)
\end{table}
%##################################################################################################

\subsection{Evaluation on Seen Identities}
We further evaluate our model's performance on seen identities, with the results shown in Table~\ref{tab:biwi_seen_id}. 
The evaluation is performed on the BIWI Test-A set. 
We compare our \textbf{MemoryTalker} with CodeTalker~\cite{codetalker2023}, a one-hot-based method, and UniTalker~\cite{unitalker2024}, the current state-of-the-art approach. 
For a fair comparison, the correct one-hot identity vector is provided to CodeTalker. 
As demonstrated in the table, our method outperforms both competing methods in the seen identity setting as well.

%------------------------------------ tableS4
%##################################################################################################
\begin{table}[!t]
\centering
\renewcommand{\arraystretch}{1.4} % 줄 간격 조정
\renewcommand{\tabcolsep}{1.5mm} % 열 간격 조정
\resizebox{\linewidth}{!}{ % 현재 컬럼 너비에 맞게 테이블 크기 조정
% 열 지정자: lcccc (Method, FVE, LVE, LDTW, Params) - LDTW 추가로 c 하나 더
\begin{tabular}{ccccc} % 표 좌우의 추가 공간 제거
\toprule
% 다음 두 줄 (VOCASET 헤더 및 cline)을 제거했습니다.
%& \multicolumn{3}{c}{VOCASET \cite{VOCA2019}} \\
%\cline{2-4} 
\multirow{2}{*}{Method} & $\text{FVE}\downarrow$ & $\text{LVE}\downarrow$ & LDTW$\downarrow$       & Params \\ % LDTW 컬럼 헤더 추가
                        & $(\times 10^{-6})$    & $(\times 10^{-5})$    &  $(\times 10^{-5})$      & (M)    \\ 
\midrule
ASR for $E_s$ & 0.518 & 0.300 & 0.437 & 185 \\ % LDTW 값 0.437 추가, Params 값 (예시 값)
\rowcolor{gray!15} % 행 배경색 (xcolor 패키지 필요)
\textbf{MemoryTalker (Ours)} & \textbf{0.506} & \textbf{0.293} & \textbf{0.418} & \textbf{94}  \\ % LDTW 값 418 추가, Params 값 (예시 값, 더 좋다면 볼드)
\bottomrule
\end{tabular}
}
\caption{Comparison the style features extracted by Mel-spectrogram- and ASR.} % 캡션에 Params 언급 추가 (캡션 내용은 필요에 따라 LDTW도 언급하도록 수정 가능)
\label{tab:vocaset_comparison_styleablation} % 레이블 수정 (선택 사항)
\end{table}
%##################################################################################################

\subsection{Analysis on Speaking Style Features}
We justify our choice of using mel-spectrograms for the speaking style encoder. 
Mel-spectrograms are well-known for preserving rich acoustic details such as prosody, rhythm, and timbre \cite{icml18}. 
In contrast, features from an Automatic Speech Recognition (ASR) model are less suitable for this task because ASR models are trained to extract neutral text representations, thereby suppressing speaker-specific identity and style. 

To empirically validate this, we present an ablation study in Table~\ref{tab:vocaset_comparison_styleablation}, where we replace our mel-spectrogram-based style encoder with an ASR-based one. 
The results show a degradation in performance, confirming that mel-spectrograms are better suited for capturing the nuances of speaking style for our task.

%------------------------------------ Figure_S2
%##################################################################################################
\begin{figure*}[t!]
\begin{center}
\includegraphics[width=0.90\linewidth]{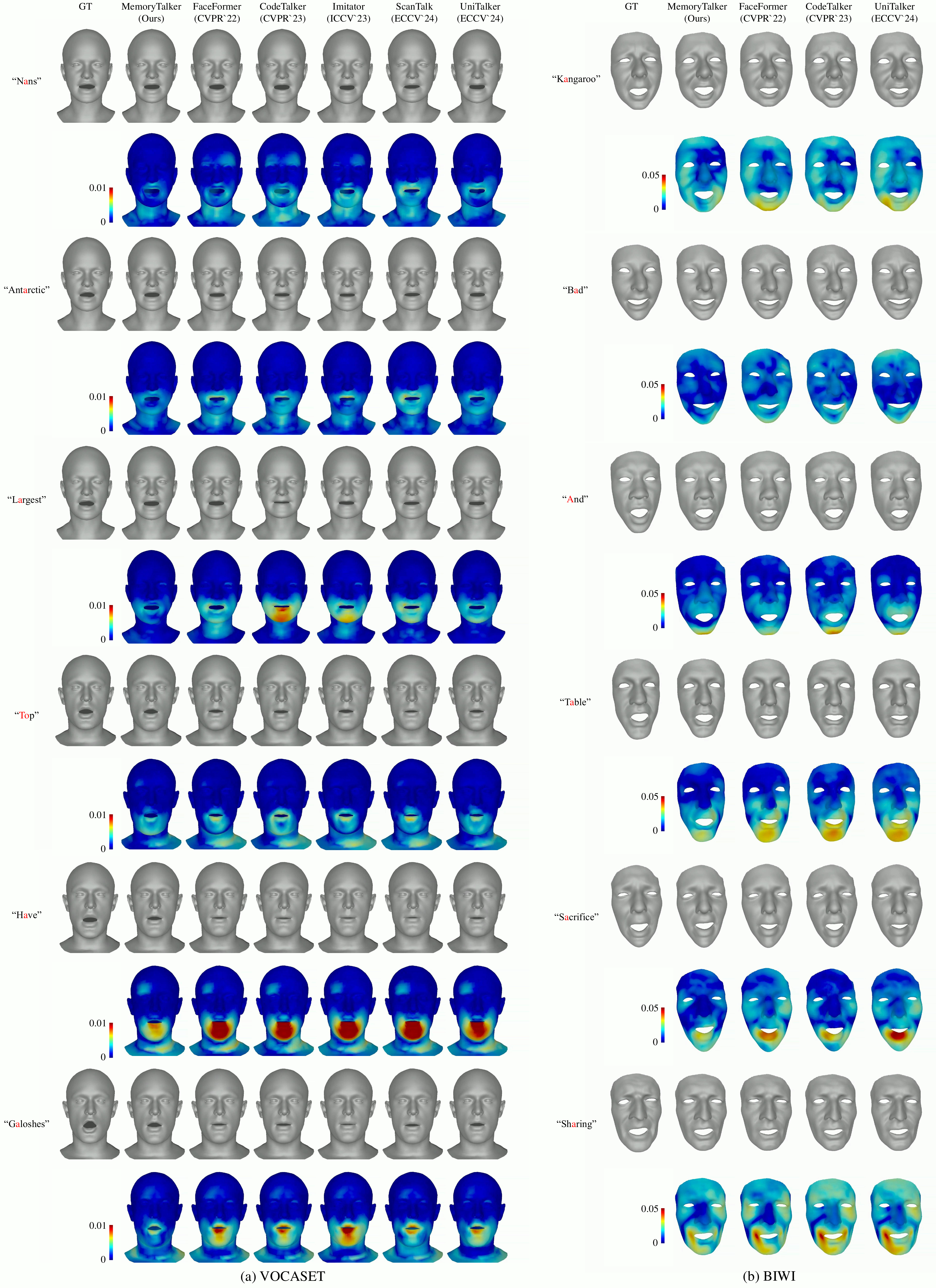}
\end{center}
    \caption{Qualitative comparisons for the cases where the pronunciation involves starting to \textcolor{red}{OPENING the mouth}.}
\vspace{-10pt}
\label{supp_fig:2}
\end{figure*}
%##################################################################################################

%------------------------------------ Figure_S3
%##################################################################################################
\begin{figure*}[t!]
\begin{center}
\includegraphics[width=0.90\linewidth]{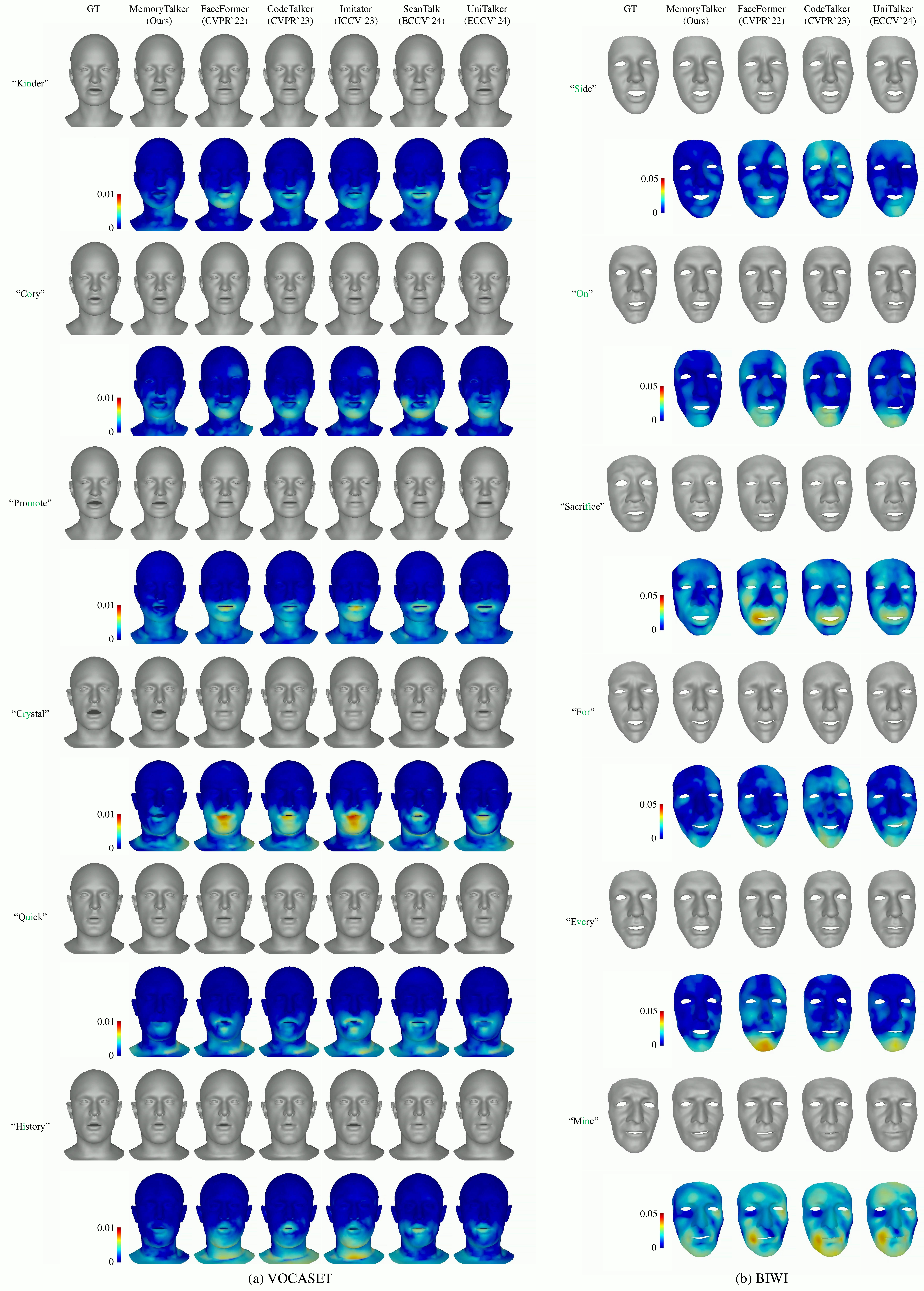}
\end{center}
    \caption{Qualitative comparisons for the cases where the pronunciation involves starting to \textcolor{darkgreen}{CLOSING the lips}.}
\vspace{-20pt}
\label{supp_fig:3}
\end{figure*}
%##################################################################################################

%------------------------------------ Figure_S4
%##################################################################################################
\begin{figure}[t!]
\begin{center}
\includegraphics[width=1.0\linewidth]{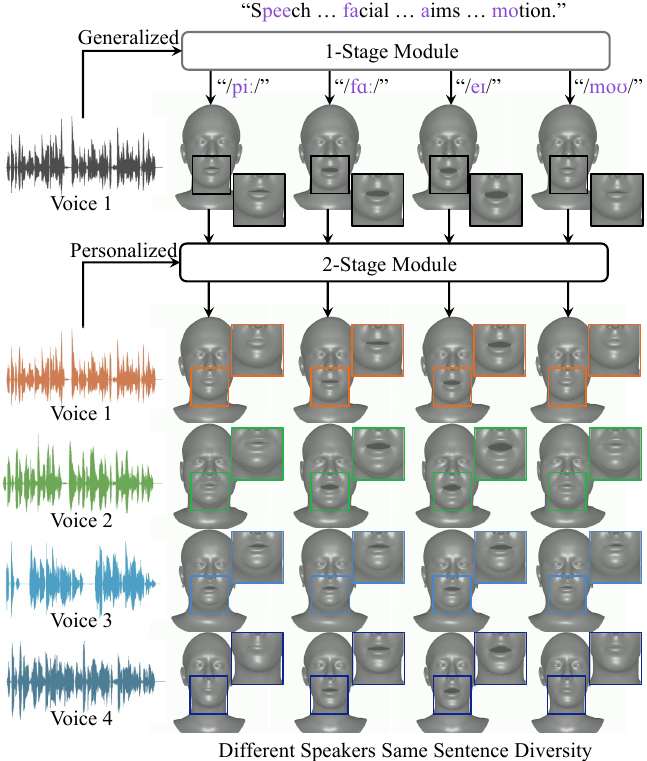}
\vspace{-0.3cm}
\end{center}
\caption
{
    % \textcolor{blue}
    {
    General Motion synchronized with audio is obtained in 1-stage (i.e. w/o 2-stage).
    Then, Personalized Motion is synthesized in 2-stage using speaking style features extracted from different speakers (i.e. w/ 2-stage).
    }
}
\label{supp_fig:various_style_adaptation}
\end{figure}
%##################################################################################################
%============================

%------------------------------------ Figure_S5
%##################################################################################################
\begin{figure}[t!]
\begin{center}
\includegraphics[width=0.8\linewidth]{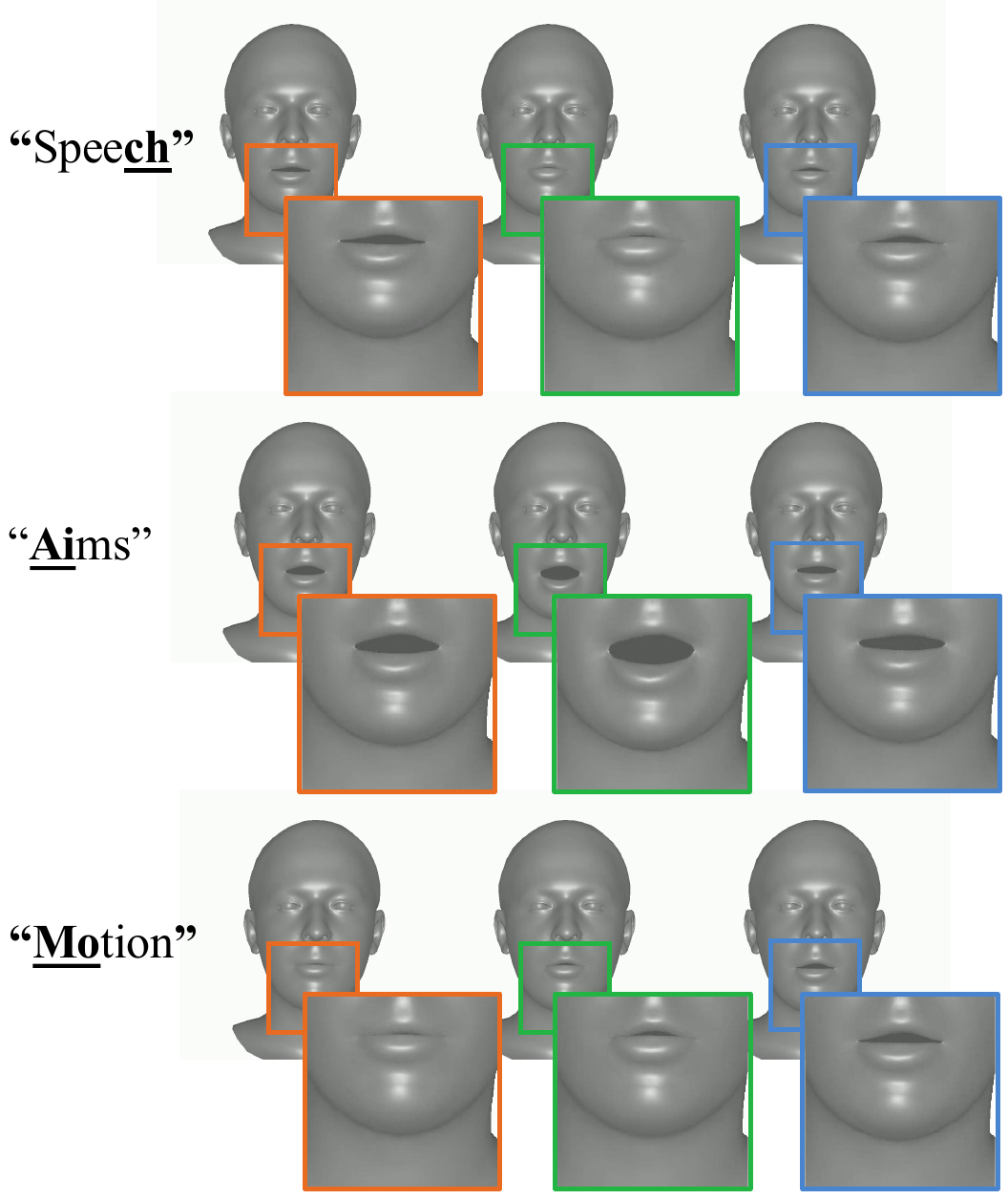}
% \vspace{-0.3cm}
\end{center}
\caption{
Results of applying different audio inputs to the same identity template. The model generates lip motions synchronized with speech while reflecting speaker-specific styles from audio, all while preserving the target’s identity.
}
\label{fig:diff_audio_same_id} % 본문 참조를 위해 레이블 수정 추천
\end{figure}
%##################################################################################################
%============================

%=========================== Qualitative Evaluation
\section{Qualitative Results}
\subsection{Performances Comparisons with Existing Methods}
Figures \ref{supp_fig:2} and \ref{supp_fig:3} show extensive qualitative results, including detailed performance comparisons with existing methods. Our analysis provides a frame-by-frame visualization, offering a more nuanced and comprehensive comparison between our proposed method and current state-of-the-art approaches. Figure \ref{supp_fig:2} illustrates cases where the pronunciation necessitates the opening of the mouth, exemplified by sounds like \textcolor{red}{/a/}. Conversely, Figure \ref{supp_fig:3} showcases instances where the pronunciation involves the initial stages of lip closure, as demonstrated by sounds such as \textcolor[rgb]{0.31, 0.69, 0.37}{/o/}. Upon careful examination of the error maps, it becomes evident that our proposed method achieves significantly more accurate results across both pronunciation types and diverse speaker profiles. This superior performance is consistently maintained when compared to all existing methods in the field. 
These results indicate that our method effectively captures and generates the respective speaking styles of multiple people under various different conditions.

%------------------------------------ Figure_S6
%##################################################################################################
\begin{figure*}[t!]
\begin{center}
\includegraphics[width=0.90\linewidth]{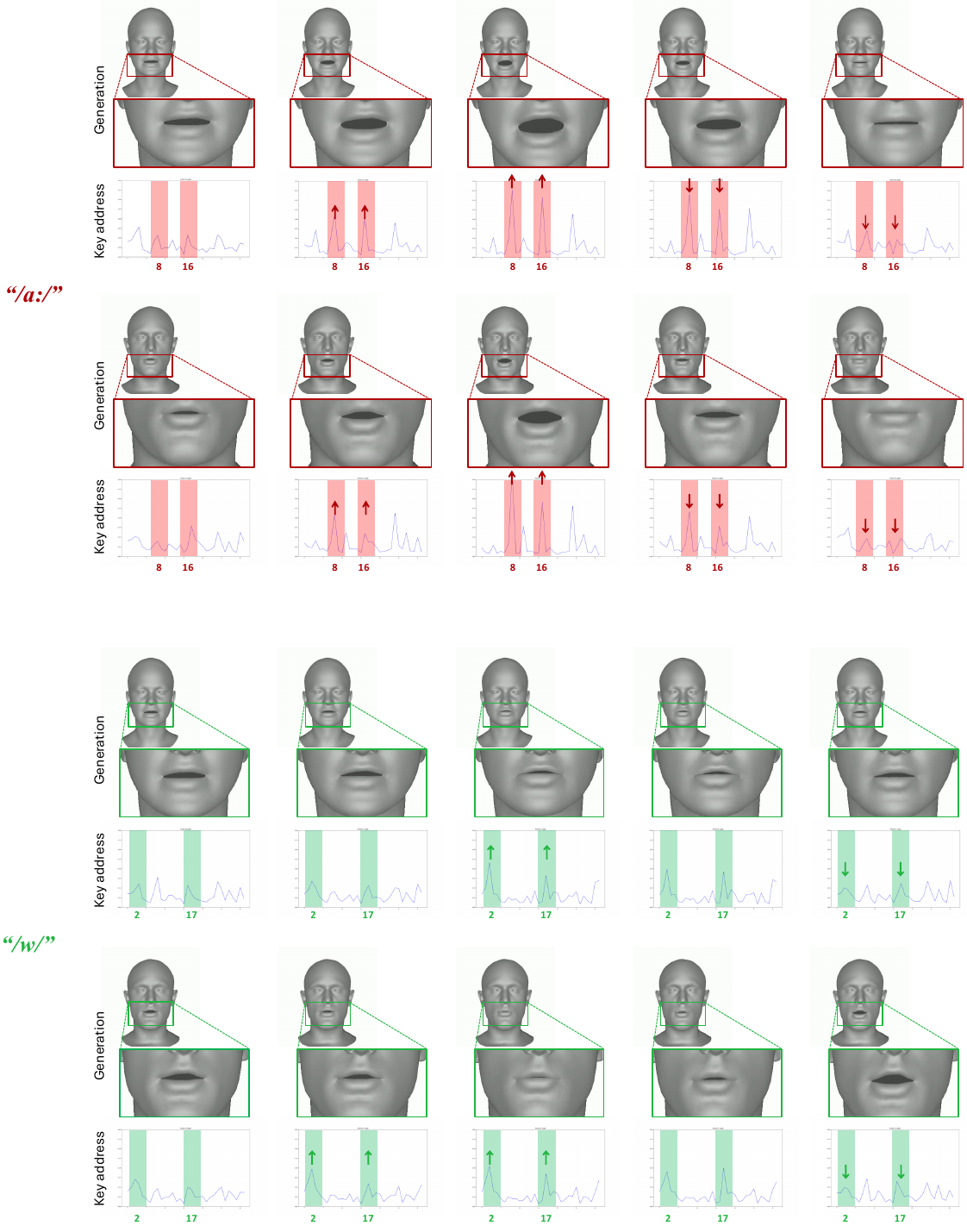}
\vspace{-0.3cm}
\end{center}
        \caption{Key addresses from audio input and corresponding generated mesh in a sequence. Note that the key addressing vector here is from the ASR model, which is to extract speaker-neutral general motions. Our personalization is further applied to the memory components afterwards.}
\label{supp_fig:key_addr}
\end{figure*}
%##################################################################################################
%============================

%------------------------------------ Figure_S7
%##################################################################################################
\begin{figure*}[t!]
\begin{center}
\includegraphics[width=0.99\linewidth]{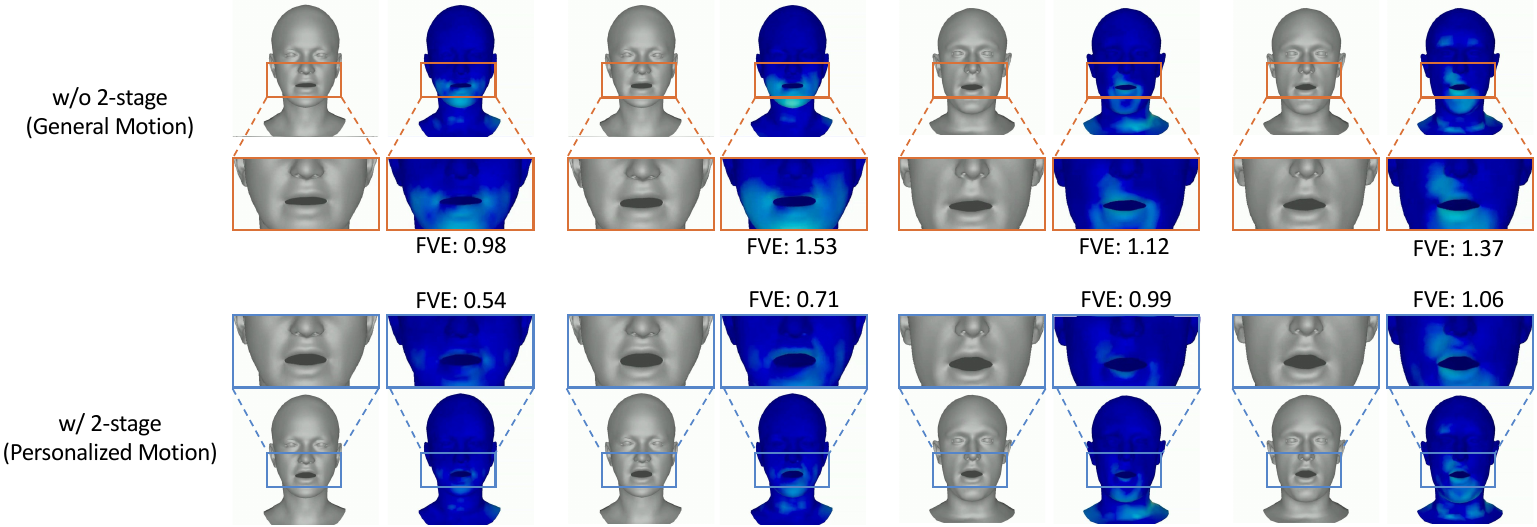}
\end{center}
    \caption{Qualitative comparison of results of learning with 1-stage only (general motion) and those of learning with 2-stage (personalized motion).}

\label{supp_fig:1vs2}
\end{figure*}
%##################################################################################################

%------------------------------------ Figure_S8
%##################################################################################################
\begin{figure*}[t!]
\begin{center}
\includegraphics[width=0.8\linewidth]{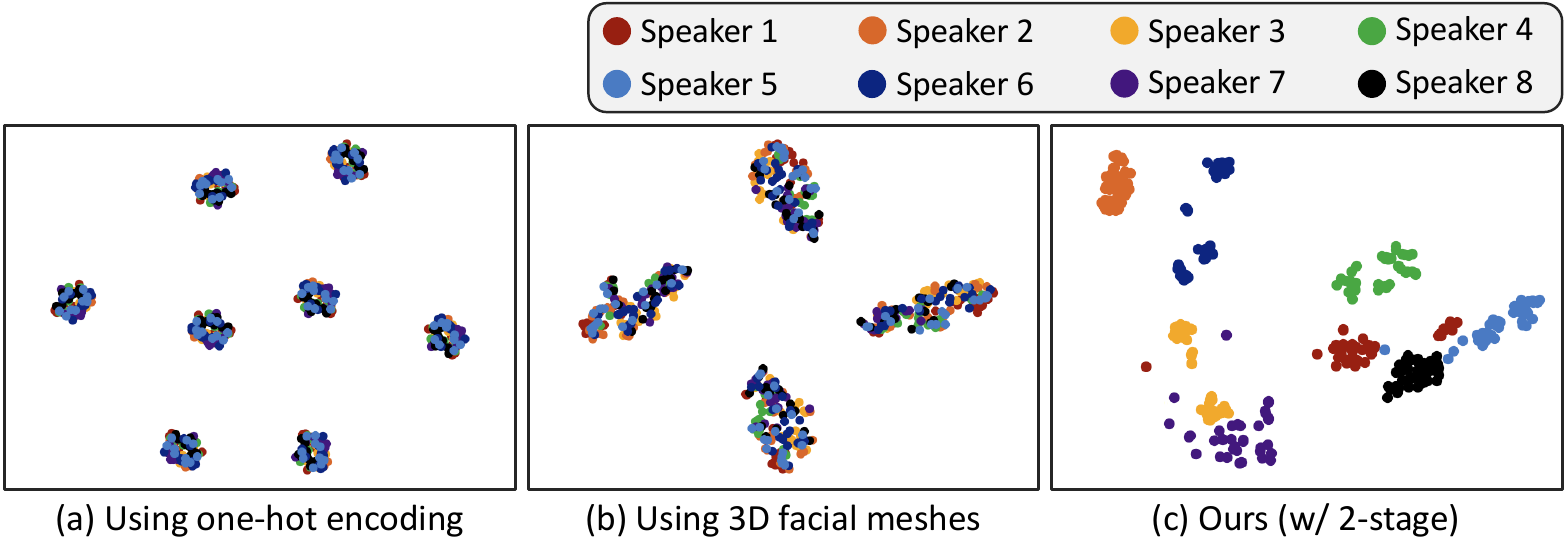}
\end{center}
        \caption{Feature visualization according to different style encoding types: one-hot, 3D facial meshes, and ours.}
\label{supp_fig:4}
% \vspace{-7pt}
\end{figure*}
%##################################################################################################
%============================

%------------------------------------ Table_S5
%##################################################################################################
\begin{table*}[ht!]
\centering
\begin{tabular}{lll}
\hline
\textbf{Module}                 & \textbf{Input→Output}                                     & \textbf{Operation} \\ \hline
\textbf{Motion Encoder}         & $\mathbf{Motion}(T, V \times 3) \rightarrow f_m(T, d_m)$         & Linear(V $\times$ 3, $d_m$) \\ \hline
\textbf{Speaker Style Encoder}  & $\mathbf{MelSpec}(T_{mel}, 80) \rightarrow \tilde{f}_s(d_{txt})$                    & \begin{tabular}[c]{@{}l@{}}
Conv1d(80, 128) $\times$ 8 → [$B$, $T_{mel}$, 1280]\\ 
Conv1d(1280, 128) → [$B$, $T_{mel}$, 128] \\ 
GroupNorm(128/16, 128)\\
Conv1d(128, 128) $\times$ 6 → [$B$, $T_{mel}$, 128] \\ 
GroupNorm(128/16, 128)\\
Conv1d(128, 128) $\times$ 6 → [$B$, $T_{mel}$/8, 128] \\ 
GroupNorm(128/16, 128)\\
AdaptiveAvgPool1d(1) → [$B$, 128] \\ 
Linear(128, 128) $\times$ 6 \\ 
Linear(128, 128) $\times$ 6 \\ 
Linear(128, 256) → [$B$, 256] \\
Linear(256, $d_{txt}$) → [$B$, $d_{txt}$] \\
\end{tabular} \\ \hline
\textbf{Motion Memory}          & ${f_{txt}(d_{txt}) \rightarrow f_m(d_m)}$                     & Parameter($d_{txt}$, $d_m$) \\ \hline
\textbf{ASR Encoder}            & $\mathbf{Audio}(T_a) \rightarrow f_{txt}(T, d_{txt})$ & \begin{tabular}[c]{@{}l@{}}
Conv1d(1, $d_{ASR}$) → [$B$, $T_a$, $d_{ASR}$] \\ 
LayerNorm($d_{ASR}$)\\
LinearInterpolation $\rightarrow$ [$B$, $T$, $d_{ASR}$]\\
Conv1d($d_{ASR}$, $d_{ASR}$) $\times$ 5 → [$B$, $T$, $d_{ASR}$] \\ 
LayerNorm($d_{ASR}$)\\
Transformer($d_{ASR}$) \\ 
Linear($d_{ASR}$, $d_{txt}$) → [$B$, $T$, $d_{txt}$]
\end{tabular} \\ \hline
\textbf{Motion Decoder}         & \begin{tabular}[c]{@{}l@{}}${f_m(T, d_m+d_{txt}) \rightarrow \mathbf{Motion}(T, V \times 3)}$ \\ \end{tabular} & 
\begin{tabular}[c]{@{}l@{}}
Linear($d_m + d_{txt}$, $d_m$) \\ 
Transformer($d_m$) \\ 
Linear($d_m$, $V$ $\times$ 3)
\end{tabular} \\ \hline
\end{tabular}
\caption{The detailed architecture of our MemoryTalker.}
\label{supp_table:2}
\end{table*}
%##################################################################################################

\subsection{The Effectiveness of Stylized Motion Memory in 2-Stage}
Figure \ref{supp_fig:1vs2} shows the effectiveness of the stylized motion memory in 2-stage. In Figure \ref{supp_fig:1vs2}, the first row shows the facial mesh and error map rendered by the \textit{general} motion feature retrieved from motion memory $\textbf{M}_m$ in 1-stage. On the other hand, the fourth row shows the facial mesh and error map rendered by the \textit{personalized} motion feature recalled from the stylized motion memory $\tilde{\textbf{M}}_m$ in 2-stage. The second and third rows show the magnified versions of the lip regions of the first and fourth rows, respectively. As shown in Figure \ref{supp_fig:1vs2}, the limited lip movement can be observed in the first and second rows (results of 1-stage only, i.e., learning without 2-stage) compared to the third and fourth rows (results of learning with 2-stage). These results show that the motion memory $\textbf{M}_m$ in 1-stage has a limitation in representing the subtle personal speaking style. On the other hand, by recalling the personalized motion feature from the stylized motion memory $\tilde{\textbf{M}}_m$ in 2-stage, the 3D facial mesh can achieve fewer errors reflecting the individual's delicate speaking style.

% \textcolor{blue}
{
\subsection{Applying Style feature to General Motion}
Figure \ref{supp_fig:various_style_adaptation} demonstrates the effectiveness of reflecting the style feature in 2-stage.
In Figure \ref{supp_fig:various_style_adaptation}, the first row shows the generated 3D facial motion from synchronized with audio at 1-stage. At this time, general motion corresponding to the audio is synthesized.
On the other hand, the second, thrid, fourth, and fifth row demonstrates when the style features generated differently for each person in the 2-stage were applied to the general motion generated in the 1-stage, the subtle changed lip shape in the 2-stage was visualized.
Through this, we verify that our method not only supplements the information lacking in general motion through speaking style features, but also effectively learns speaking style information for each speaker.

% cam-ready 추가================
To demonstrate that our model performs audio-driven stylization rather than merely mirroring a fixed template, we applied different audio clips to the same identity. As shown in Figure~\ref{fig:diff_audio_same_id}, the generated animations not only produce lip motions that are accurately synchronized with each speech input but also reflect speaker-specific styles embedded in the audio—such as variations in pronunciation strength and articulation—while consistently preserving the target pronunciation.
% cam-ready 추가================

%=========================== Visualization
\section{Feature Analysis}
\subsection{Speaking Style Feature Visualization}
We provide visualizations of motion features with t-SNE \cite{t-SNE08} to show that our method reflects speaking styles in the audio, contrasting it with existing approaches (see Figure \ref{supp_fig:4}). Figure \ref{supp_fig:4} (a) visualizes the motion feature synthesized during inference when using one-hot encoding \cite{faceformer2022}. As in \cite{faceformer2022}, since there are not able to know one-hot class information, one-hot vectors are arbitrarily selected (8 classes). As a result, this model cannot distinguish actual unseen speakers at all. Figure \ref{supp_fig:4} (b) shows the results of utilizing 3D facial mesh sequence \cite{mimic2024}. It considers 3D facial mesh sequences as additional inputs. However, the 3D facial meshes are usually unavailable in real-world situations at inference. In addition, it does not distinguish the style distribution corresponding to the unseen speakers. On the other hand, as shown in Figure \ref{supp_fig:4} (c), our model provides clear speakers' clustering that corresponds to individual speaking styles. In particular, the proposed method does not require any prior information (\textit{i.e.}, speaker information) in both training and inference stages, which makes it more practical in real-world scenarios.

\subsection{Key Addressing Vector Visualization}
To verify which key addressing vector correctly retrieves the motion memory, we visualize the key addressing vectors of our model. 
Figure \ref{supp_fig:key_addr} shows the generated 3D facial mesh sequences and the corresponding key addressing vectors of MemoryTalker model. The key addressing vectors are generated from an audio segment pertaining to the phoneme \textcolor{red}{``/a:/''} and \textcolor[rgb]{0.31, 0.69, 0.37}{``/w/''}, respectively, for different speakers. 
We can see that the address smoothly varies as the lip region in the mesh moves. 
From visualizing the key addressing vectors of different speakers speaking the same pronunciation, we observe the similar tendency of the key addressing vectors.
Focusing on the slots that noticeably change their address value, from the 3rd column, the address on the 8th and 16th slot addresses increases from 0.075 to 0.200 when pronouncing \textcolor{red}{``/a:/''}, and also when presented with other speakers' speech. 
Similarly, when pronouncing \textcolor[rgb]{0.31, 0.69, 0.37}{``/w/''}, it shows that the address on the 2nd and 17th slot addresses increases regardless of speaker. 
These demonstrate that the key addressing vectors of our MemoryTalker are activated in the same slot when synthesizing the same lip shape, suggesting that our motion memory slot feature accurately stores the corresponding lip motion. Note that the key addressing vector here is from the ASR model, which is to extract speaker-neutral general motions. Our personalization is further applied to the memory components, not the key addressing vector.

%============================

%=========================== Implementation Details
\section{Implementation Details}

\subsection{Details of MemoryTalker Architecture}
We configured the main models as in Table \ref{supp_table:2}. This is the baseline architecture used in our all experiments. The Motion Encoder is designed as a single linear layer as in \cite{faceformer2022, codetalker2023, imitator2023, mimic2024}.
First, the Speaking Style Encoder concatenates the output and input features of each layer to create a 1280-dimensional feature. After that, it goes through conv1d, group normalization, and ReLU in order, and then learns the features that appear throughout the frame through an average pooling layer. Finally, it is projected as a feature with a dimension of $d_{txt}$ through a linear layer. \cite{interspeech22}
The Motion Memory uses a text addressing vector with a dimension of $d_{txt}$ as a query and emits a motion feature with a dimension of $d_m$.
The ASR encoder uses the HuBERT \cite{HuBERT2021} structure and learned parameters. First, the audio feature is encoded with a 1D convolution layer and a group normalization layer. The GeLU nonlinear function is used during encoding. After that, linear interpolation is used to match the audio fps with the motion fps (30 fps for VOCASET and 25 fps for BIWI). After that, it passes through a transformer encoder, and then a pre-trained linear layer is used to map the feature to the vocab size. The Motion Decoder reduces the dimension to $d_m$ using a linear layer, combining the retrieved motion feature and the text representation feature, and performs positional encoding. After that, the transformer decoder is used to induce the motion feature that matches the text representation. The obtained motion feature passes through a linear layer and synthesizes a motion that moves a neutral face mesh.

% \textcolor{blue}{Furthermore, we used a baseline model for ablation study about each modules. This model consists of audio encoder, transformer layer, and motion decoder used in above mentioned structure. First The audio is encoded by audio encoder, and then audio feature, which is encoded by audio encoder and applied Positional encoding, are guided to generate motion features by referencing the audio in that frame by the Transformer's self-attention mechanism. At this time, To refer the audio in that frame only, we use mask. Finally, we gain the motion synced with audio by decoding the motion features go through single linear layer used in a motion decoder.}

\subsection{Detail of User Study}
In Figure \ref{supp_fig:UserStudy}, we attach the user study we provided to the subjects.
We evaluated our method against other methods using the user study form used in the faceformer \cite{faceformer2022} and the mimic \cite{mimic2024}.
Our qualitative evaluation questionnaire consists of a total of {90} questions. 
{Five} questions are to check whether the participant is participating sincerely through the qualification video and to remove outliers. 
For the remaining {85} questions, we evaluate three qualitative metrics in total to compare the output of our \textit{MemoryTalker} with the outputs of existing SoTA methods and GT for sentences randomly sampled from the 40 sentences.

% \textcolor{blue}
{
For the evaluation of Realism, we induce the participant to choose A/B pairs by asking the following question: "Comparing the two full faces, which one looks more realistic?". In this case, participants see the full face in two samples and choose the more natural option.
}

% \textcolor{blue}
{
To measure the Lip-sync, we conducted an A/B test with the question: "Comparing the two lips, which one looks more realistic?". As with evaluating realism, we asked participants to choose the sample that was more synchronized with the audio among the A/B samples. This allowed us to subjectively evaluate the degree of synchronization between the audio and the lip region.
}
% \textcolor{blue}
{
To measure how well our \textit{MemoryTalker} captures the speaking style of the ground truth (GT), we compare whether our output is more similar to the GT than the outputs from other models.
To this end, we randomly placed the GT video in the first position, and the videos to be compared in the second and third positions.
And then, we asked the participant to answer the question "Comparing the speaking style (including the amplitude of mouth opening and closing, the dimensionality of pouting, etc.) of the last two faces, which one is more consistent with the first video?". 
}

% \textcolor{blue}
{
Additionally, to assess participants’ reliability in the qualification question, we randomly place the same first-position video in either the second or the third position.
At this time, the participant is asked a question about speaking style ("Comparing the speaking style (including the amplitude of mouth opening and closing, the dimensionality of pouting, etc.) of the last two faces, which one is more consistent with the first video?"), and this choice is different from the previous one in that it has a fixed correct answer.
Therefore, if the participant answers this question incorrectly, we consider the participant an outlier and remove it from the statistics, thereby improving the quality of our evaluation method. 
}

%------------------------------------ Figure_S9
%##################################################################################################
\begin{figure*}[t!]
\begin{center}
\includegraphics[width=0.99\linewidth]{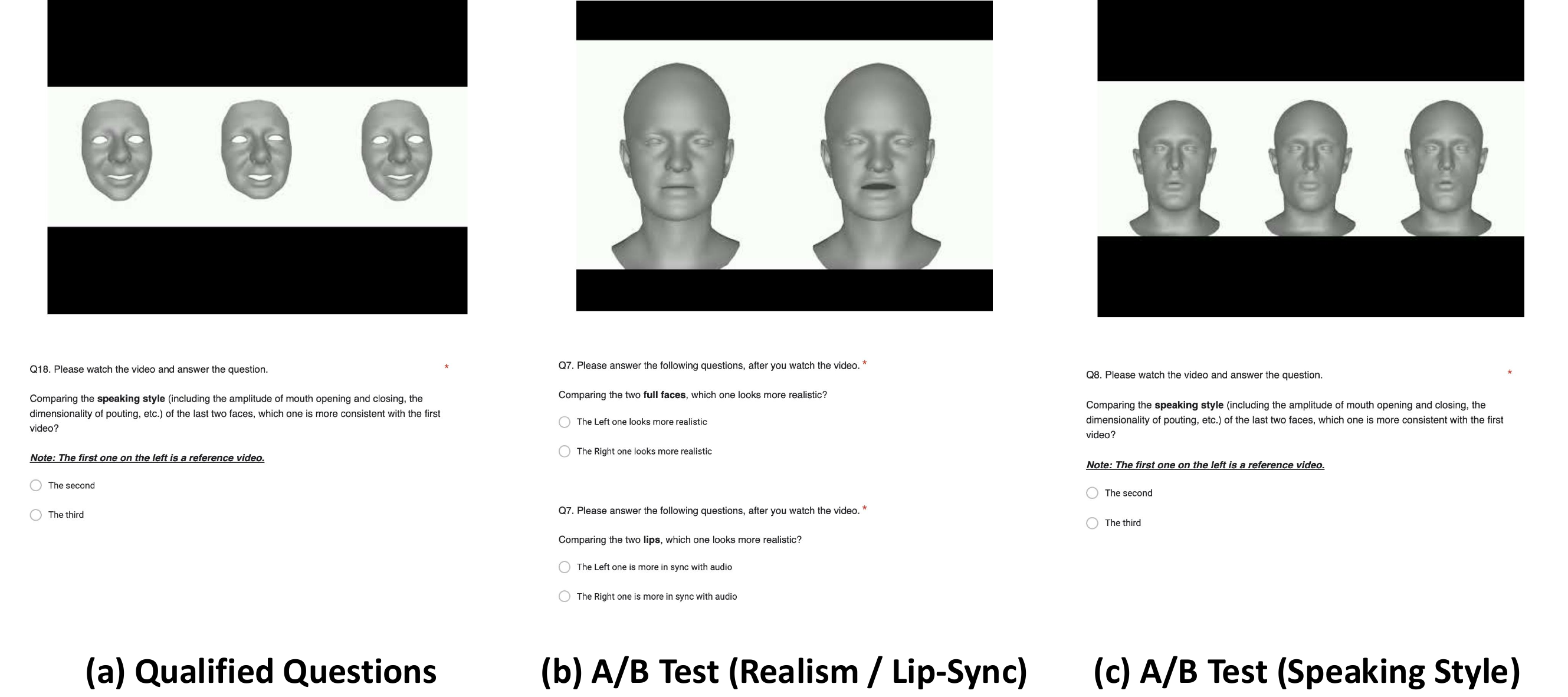}
\end{center}
    \caption{Examples of conducted user study.}
% \vspace{-10pt}
\label{supp_fig:UserStudy}
\end{figure*}
%##################################################################################################

%------------------------------------ Table_S6
%##################################################################################################

\begin{table}[!t]
\renewcommand{\arraystretch}{0.8}
\centering
\small % 또는 \footnotesize 사용 가능
\resizebox{0.8\linewidth}{!}{% % 표 크기를 줄여 글씨 크기도 줄임
\begin{tabular}{lcc}
\toprule
{Competitors}                           & Lip Sync(\%)  & Realism(\%)       \\
\midrule
vs. GT                                  & 41.1           & 40.3               \\
\bottomrule
\end{tabular}
}
\caption{User study: our method vs. GT on  VOCASET \cite{VOCA2019}.}
\label{table:supp_userstudy}
% \vspace{-7pt}
\end{table}
%##################################################################################################

% \textcolor{blue}
{
In summary, we evaluated three qualitative metrics (Speaking Style, Realism, and Lip-sync) by randomly selecting ten sentences per model, following a similar scale to previous study\cite{faceformer2022}. Based on the ten sampled videos for each model, we split them into two groups: five videos were used to evaluate Speaking Style, and the remaining five were used to evaluate Realism and Lip-sync. Specifically, we generated one question per video for Speaking Style (5 questions), and two questions per video (one for Realism, one for Lip-sync) for the other five videos (10 questions). Consequently, each model yielded a total of 15 questions (5 + 10) from its ten samples. With five models, we obtained 50 samples in total and generated 90 questions.
Additionally, we selected ten more samples to directly compare our method with the ground truth (GT) as \ref{table:supp_userstudy}. Among these, five were used for qualification questions, and the remaining five were used for Realism and Lip-sync evaluations between our approach and the GT. All questions were administered in a forced-choice format, and any participant who answered even one qualification question incorrectly was excluded from the final statistics.
Furthermore, we provided two reminders to participants: (1) to ensure their computer’s sound was on while watching the videos, and (2) to note that one or two of the videos might be used for qualification purposes, such that random guessing could result in disqualification. Because each video is relatively short (4–7 seconds), we allowed participants to watch them repeatedly up to five times and even replay them additionally, ensuring they could carefully assess each sample.
}
A total of {33} people participated in our user study, and 2 of them were removed because they did not pass the qualification questions we provided. In our experience, the participants took about 40 minutes to complete our user study.

\end{document}